\documentclass{article} 
\usepackage{iclr2025_conference,times}

\usepackage{amsmath,amsfonts,bm}

\def\eqref#1{equation~\ref{#1}}

\def\1{\bm{1}}

\DeclareMathAlphabet{\mathsfit}{\encodingdefault}{\sfdefault}{m}{sl}
\SetMathAlphabet{\mathsfit}{bold}{\encodingdefault}{\sfdefault}{bx}{n}

\usepackage{natbib}
\PassOptionsToPackage{table}{xcolor}
\usepackage{hyperref}
\usepackage{url}
\usepackage{booktabs}
\usepackage{graphicx}
\usepackage[ruled,vlined]{algorithm2e}
\usepackage{algorithmic}
\usepackage{amsmath} 
\usepackage{amsthm} 
\usepackage{tikz}
\usepackage{multirow}
\usepackage{xcolor}
\usepackage{colortbl}
\usepackage{amssymb}
\usepackage{pifont} 
\usepackage{makecell}
\definecolor{lightblue}{rgb}{0.1, 0.1, 0.9}

\title{SageAttention: Accurate 8-bit attention for Plug-and-Play Inference Acceleration}

\author{Jintao Zhang, Jia Wei, Haofeng Huang, Pengle Zhang, Jun Zhu, Jianfei Chen\thanks{Corresponding author.} \\
Dept. of Comp. Sci. \& Tech., Institute for AI, BNRist Center,\\
Tsinghua-Bosch Joint ML Center, THBI Lab, Tsinghua University\\
\texttt{\{zhang-jt24@mails., jianfeic@, dcszj@\}tsinghua.edu.cn} }

\newcommand{\annotate}[1]{\textcolor{gray}{{#1}}\xspace}

\newcommand{\our}{\texttt{SageAttention}\xspace}
\newcommand{\ourt}{\texttt{SAGEAttn-T}\xspace}
\newcommand{\ourb}{\texttt{SAGEAttn-B}\xspace}
\newcommand{\ourvt}{\texttt{SAGEAttn-vT}\xspace}
\newcommand{\ourvb}{\texttt{SAGEAttn-vB}\xspace}
\newcommand{\cogvideo}{\texttt{CogvideoX}\xspace}
\newcommand{\llama}{\texttt{Llama2}\xspace}
\newcommand{\ultrapixel}{\texttt{UltraPixel}\xspace}
\newcommand{\unidiffuser}{\texttt{Unidiffuser}\xspace}
\newcommand{\timm}{\texttt{TIMM}\xspace}
\newcommand{\llava}{\texttt{Llava1.6}\xspace}

\newcommand{\purple}[1]{\textcolor{black}{{#1}}\xspace}
\newcommand{\jt}[1]{\textcolor{blue}{{#1}}\xspace}

\newcommand{\rowmax}{\mathrm{rowmax}}
\newcommand{\diag}[1]{\mathrm{diag}\left(#1\right)}
\newcommand{\mean}{\mathrm{mean}}

\definecolor{deepgreen}{rgb}{0.0, 0.5, 0.0}  
\definecolor{deepred}{rgb}{0.6, 0.0, 0.0}  

\newcommand{\dgreen}[1]{\textcolor{deepgreen}{#1}}
\newcommand{\dred}[1]{\textcolor{deepred}{#1}}

\iclrfinalcopy 
\begin{document}

\maketitle

\begin{abstract}
The transformer architecture predominates across various models. As the heart of the transformer, attention has a computational complexity of $O(N^2)$, compared to $O(N)$ for linear transformations. When handling large sequence lengths, attention becomes the primary time-consuming component. Although quantization has proven to be an effective method for accelerating model inference, existing quantization methods primarily focus on optimizing the linear layer.
In response, we first analyze the feasibility of quantization in attention detailedly. Following that, we propose \our, a highly efficient and accurate quantization method for attention. The OPS (operations per second) of our approach outperforms FlashAttention2 and xformers by about \textbf{2.1x} and \textbf{2.7x}, respectively. \our also achieves superior accuracy performance over FlashAttention3. Comprehensive experiments confirm that our approach incurs almost \textbf{no end-to-end metrics loss across diverse models}—including those for large language processing, image generation, and video generation. The code is available at \jt{\url{https://github.com/thu-ml/SageAttention}}.

\end{abstract}

\begin{figure}[h]
    \begin{center}
    \vspace{-1em}
    \includegraphics[width=0.99\textwidth]{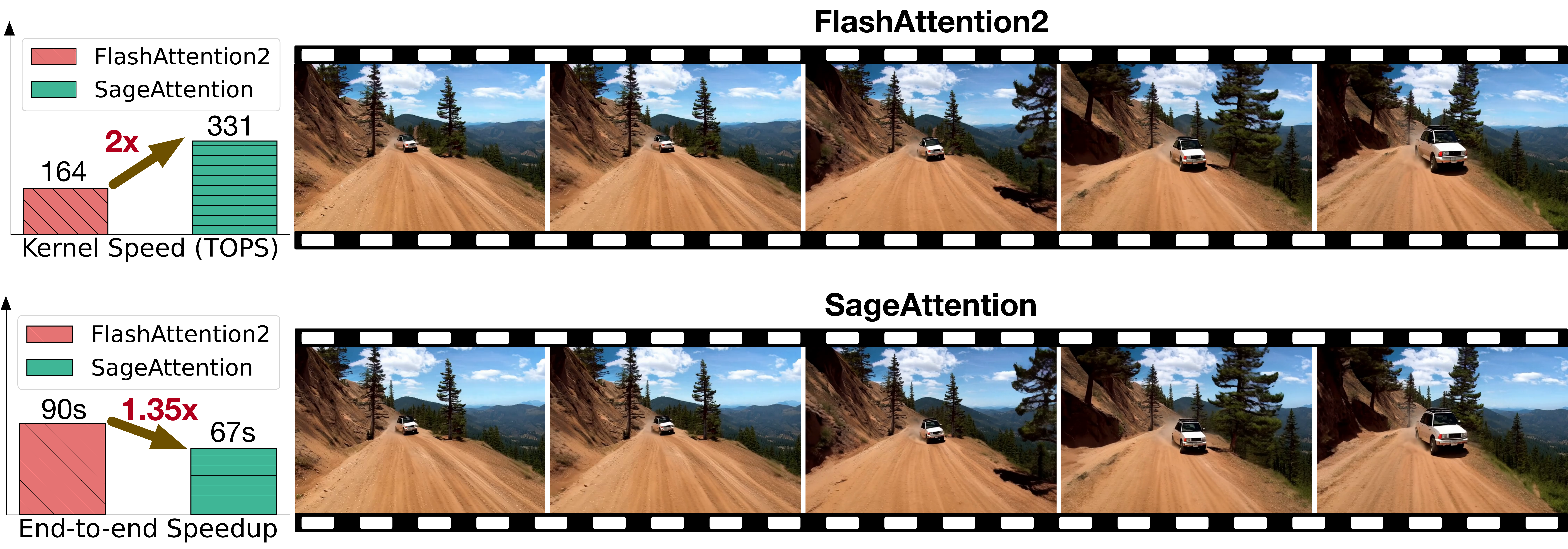}
    \vspace{-.6em}
    \caption{An example of \our on video generation (CogvideoX on RTX4090).}
    \vspace{-.6em}
    \end{center}
    \label{fig:intro}
\end{figure}

\section{Introduction}

\begin{figure}[h]
    \centering 
    \begin{minipage}[b]{0.42\linewidth} 
        \centering
        \vspace{-.2em}
        \includegraphics[width=0.93\linewidth]{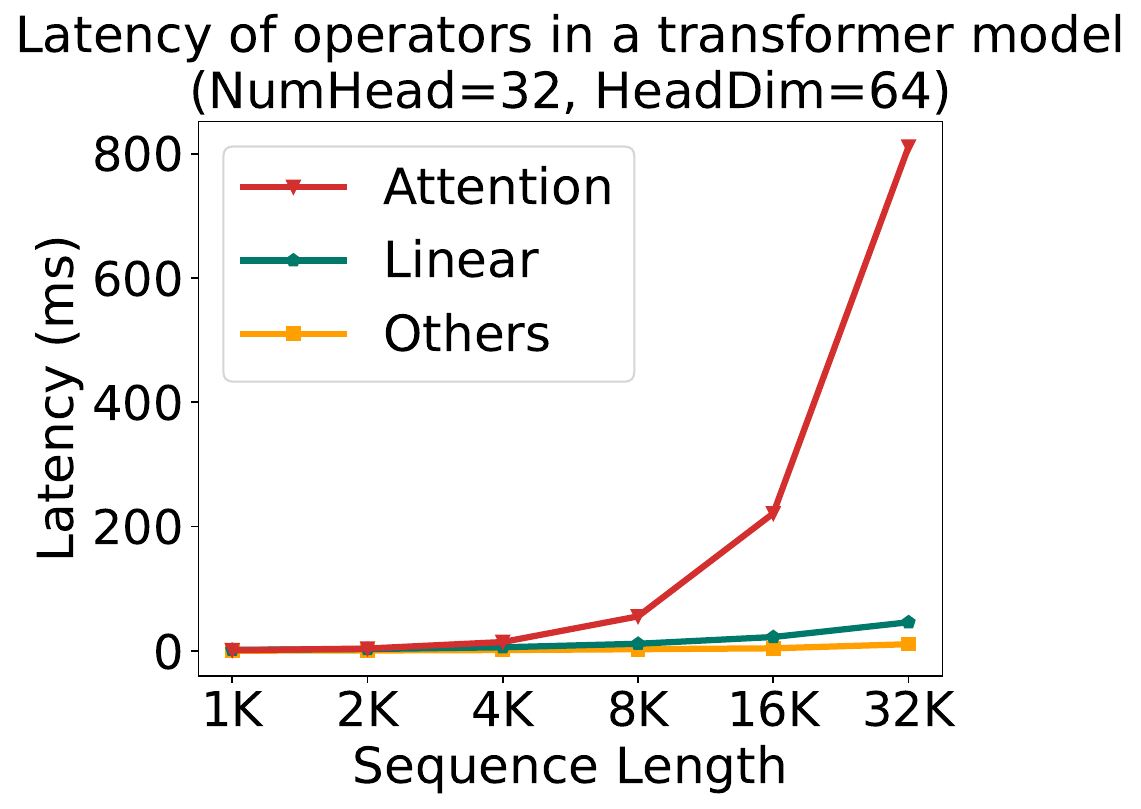} 
        \vspace{-.7em}
        \caption{Latency of attention.}
        \label{fig:motivation}
    \end{minipage}
    \hspace{0.05\textwidth} 
    \begin{minipage}[b]{0.37\linewidth}
        \centering
        \vspace{-.2em}
        \includegraphics[width=\linewidth]{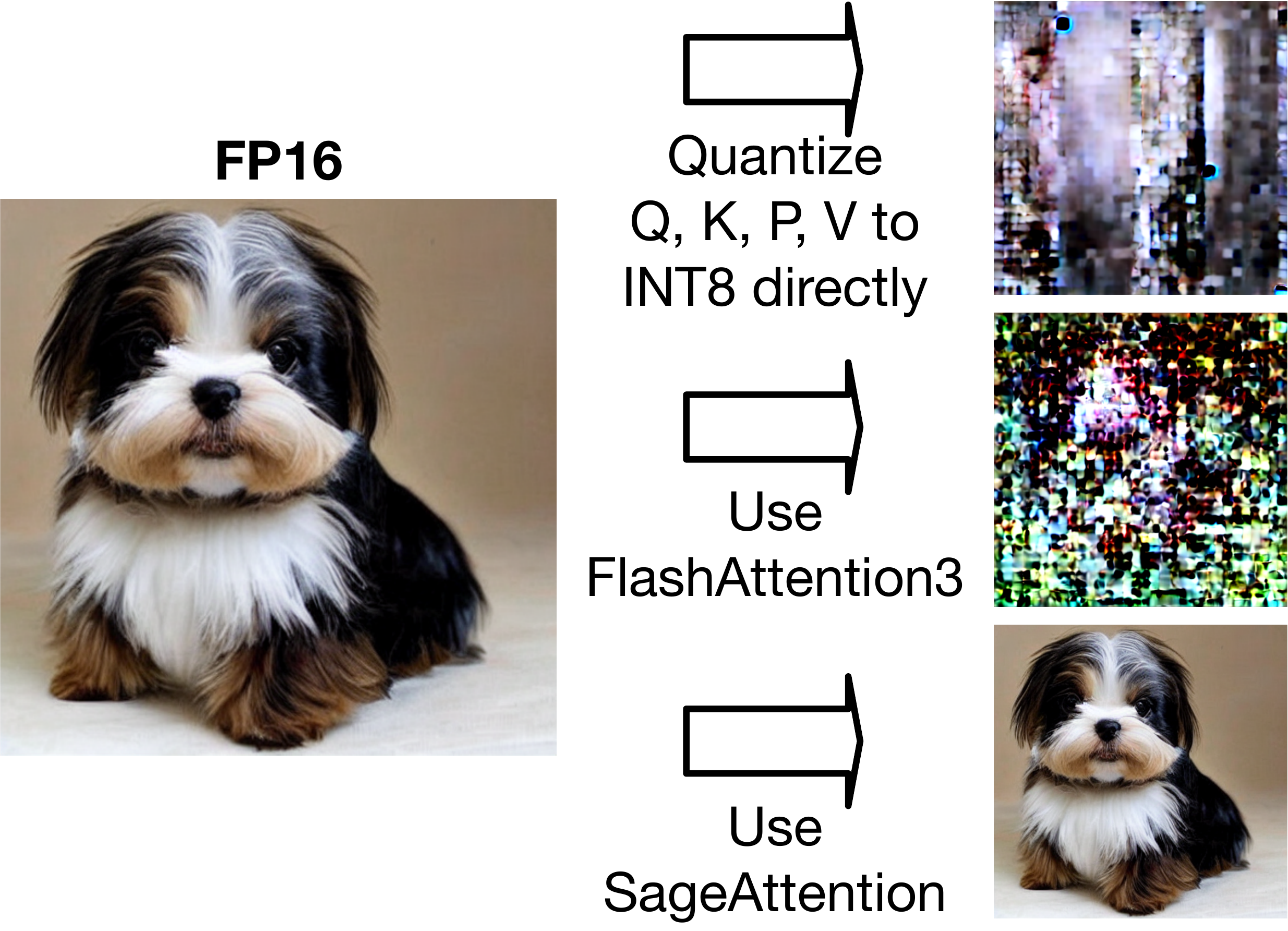} 
        \vspace{-1.7em}
        \caption{A comparison example.}
        \label{fig:challenge}
    \end{minipage}
\vspace{-1.3em}
\end{figure}

Attention is the fundamental component of transformers~\citep{vaswani2017attention}, and efficiently computing attention is crucial for transformer-based applications. 
Moreover, there is a recent trend in processing longer sequences, which further strengthens the need for faster attention. In tasks like video generation~\citep{yang2024cogvideox} and language model prefilling~\citep{llama31model}, the sequence length can easily go up to 8K$\sim$ 128K. Due to its quadratic complexity, the cost of attention dominates all other operations in such scenarios, as illustrated in Figure~\ref{fig:motivation}.

Quantization is an effective strategy for enhancing neural networks' computational and memory efficiency by reducing the numerical precision. There are abundant works on accelerating training~\citep{sun2019hybrid,xijetfire,peng2023fp8} and inference~\citep{jacob2018quantization,xiao2023smoothquant} with low-precision numerical formats such as FP8, INT8, or INT4. However, existing works primarily focused on quantizing the \emph{linear} layer, where \emph{attention} is left unaccelerated in high-precision, such as FP16. There is not yet a work that systematically investigates the quantization of attention. 
Moreover, many quantization methods require extra training, and the cost can be prohibitive for large-scale models. 
While \purple{FlashAttention3}~\citep{shah2024flashattention} was released recently and offers an FP8 version, it is \purple{tailored to and can only be used with the Nvidia Hopper architecture}. This exclusive optimization limits its broader applicability. Furthermore, our analysis demonstrates that directly implementing the FP8 version can lead to performance degradation, as detailed in Table~\ref{tab:quant_type_analysis}.

Quantizing attention is challenging. The computation of attention is more complex than that of linear operations. Attention includes a softmax operation and two matrix multiplication (Matmul) operations: $QK^\top$ and $PV$. Direct 8-bit quantization and dequantization of the matrices $(Q, K, P, V)$ in attention will result in significantly degraded performance across various models. For example, the text-to-image model Unidiffuser~\citep{bao2022uni} will generate a completely blurry image with both INT8 and FlashAttention3's FP8 implementation (See Figure~\ref{fig:challenge}), and Llama2 only achieves a random-guessing-level accuracy of 25.5\% on the MMLU dataset with INT8 attention. After investigating deeply, we identified two primary challenges: 
\textbf{(C1)} The matrix $K$ exhibits a significant channel-wise outlier, leading to substantial accuracy loss during quantization.
\textbf{(C2)} Simply quantizing $(P, V)$ into INT8 does not consistently ensure the accuracy of $PV$ across various scenarios.

In this paper, we propose \our, a quantization method to accelerate attention while \emph{preserving accuracy}. \our is easy-to-use. As a post-training quantization method, it can be \emph{used in a plug-and-play manner in inference time} by simply replacing the original high-precision implementation. We propose several techniques to achieve this goal. 
First, we opt to quantize the tensors in attention to INT8 rather than FP8. This decision is based on the fact that INT8 Matmul on some commonly used GPUs, e.g., RTX4090 and 3090, are four times faster than in FP16 and two times faster than FP8. Moreover, INT8 quantization for matrices $(Q, K)$ is more precise than FP8 in attention (See Table~\ref{tab:quant_precision_analysis1}).
To address \textbf{(C1)}, we propose a method to smooth the $K$ matrix. This method significantly enhances accuracy with a negligible time overhead ($<$0.2\%). 
To address \textbf{(C2)}, as an alternative to quantizing $(P, V)$ to 8-bit, we propose a more accurate yet efficient method for the Matmul $PV$: we maintain $(P, V)$ in FP16 and use a low-precision FP16 accumulator. This strategy doubles Matmul's speed without sacrificing any accuracy.
Finally, we implement several versions of attention with different speed-accuracy tradeoffs and propose a method to select the fastest attention implementation for each layer while preserving accuracy. 

We offer a high-performance implementation of \our on RTX4090 and 3090 GPUs using Triton~\citep{openaitriton}. Our implementation contains a fused kernel combining ROPE with quantization and a fast self-attention kernel inspired by FlashAttention-style tiling. The implementation utilizes the fast INT8 \emph{mma(u8.u8.s32)} and FP16-with-FP16-accumulator \emph{mma(f16.f16.f16)} instructions of Nvidia Tensor Core.
Our kernel is about $2.1 \times$ and $2.7 \times$ faster than FlashAttention2 and xformers, respectively.
Notably, it achieves 340 TOPS on RTX4090 at headdim=64 and headdim=128, reaching 52\% of the theoretical INT8 throughput. In contrast, the peak for the state-of-the-art FlashAttention2 is only 165 TOPS. Moreover, at headdim=64, our throughput on RTX 4090 is even close to the 490 TOPS throughput of FlashAttention3, which is exclusive to the much more powerful and expensive Hopper GPUs.
We extensively evaluate the end-to-end metrics of our approach on state-of-the-art image/video generation, image classification, and language models. On all tasks, \our can be directly adopted in a plug-and-play manner with negligible loss in model performance, while offering more than 2$\times$ speedup than FlashAttention2 and xformers.

\section{Related Work}

We categorize efficient Attention works into three groups: 
\textbf{(1) Sparse Attention.} This strategy only selects parts of a sequence from a given context for processing with standard Attention. Implementations like Swin transformer~\citep{liu2021swin}, Twins~\citep{chu2021twins}, UniFormer~\citep{liuniformer}, Attentionsinks~\citep{xiao2023efficient}, InfLLM~\citep{xiao2024infllm}, LongLora~\citep{chen2023longlora}, Minference~\citep{jiang2024minference}, and SkipAttention~\citep{venkataramanan2023skip} show promise. However, these methods' limitations are that they only work in a few scenarios because omitted calculations are not always useless.
\textbf{(2) Linear Attention.} Techniques that transform Attention computation to reduce time complexity, for example, Linformer~\citep{wang2020linformer}, Performer~\citep{choromanski2020rethinking}, MetaFormer~\citep{yu2022metaformer}, and LinearAttention~\citep{katharopoulos2020transformers}, which lower the time complexity of Attention from $O(N^2)$ into $O(N)$. These methods excel in specific scenarios while standard Attention remains prevalent.
\textbf{(3) Kernel Optimization.} Rather than simplifying calculations, these methods exploit hardware capacities to enhance speed. The xformers~\citep{xFormers2022} platform accelerates Attention with customizable blocks and dedicated CUDA kernels. FlashAttention~\citep{dao2022flashattention} proposes tiling to reduce the memory reads/writes between GPU global memory and on-chip SRAM for significant speedups. FlashAttention2~\citep{dao2023flashattention} refine the parallelism and warps partition of FlashAttention. \citet{bikshandi2023case} further optimize FlashAttention2 by kernel fusion. FlashAttention3~\citep{shah2024flashattention} is proposed for Hopper architecture. However, FlashAttention3 is exclusive to the Hopper GPU architecture, and the accuracy of its quantization version is significantly lower than our method (See Table~\ref{tab:quant_type_analysis}). RingAttention~\citep{liuringattention} scales FlashAttention across multiple GPUs. I-bert~\citep{kim2021bert} quantizes all tensors in a transformer block into INT8 but is restricted to RoBERTa.
Our method falls under the third category, and is \underline{\textbf{orthotopic}} with the first and second categories.

\section{Preliminary}

Our method builds on FlashAttention-2 and adopts dynamic quantization.  We will begin by reviewing FlashAttention-2, followed by a brief introduction to dynamic quantization techniques.

\subsection{FlashAttention} \label{sec:flash_attn}

The computation of self-attention can be formulated as follows: $S = Q K^\top/\sqrt{d},~P = \sigma(S),~O = P V$, where $\sigma(S)_{ij} = \exp(S_{ij})/\sum_{k} \exp(S_{ik})$ is the softmax operation.
The matrices $Q$, $K$, and $V$ each have dimensions $N \times d$, while the matrices $S$, $P$ are $N \times N$. While $d$ is typically small, e.g., 64 or 128, $N$ can be thousands if not millions. Therefore, the $N\times N$ matrices $(S, P)$ are much larger than $(Q, K, V)$, and a naive implementation suffers from the huge amount of global memory I/O for $(S, P)$ reads/writes. FlashAttention~\citep{dao2023flashattention} proposes to tile $Q$, $K$, and $V$ from the token dimension into blocks $\{Q_i\}, \{K_i\}, \{V_i\}$ with block sizes of $b_q$, $b_{kv}$, $b_{kv}$, respectively. Then, to avoid the memory I/O for $(S, P)$, it uses online softmax~\citep{milakov2018online} to progressively compute each block of $O$, i.e., $O_i$ as follows.

First, for each block of $\{K_i\}, \{V_i\}$, it computes the following equations iteratively:

\begin{align}
 S^j_i = Q_i K_j^\top / \sqrt{d},& ~~~(m^{j}_i, \widetilde P^j_i) = \tilde\sigma(m^{j-1}_i, S^j_i),  \\
 l_i^j = \exp(m^{\purple{j-1}}_i-m^{\purple{j}}_i) l_i^{j-1} + \mathrm{rowsum}(\widetilde P^j_i),& ~~O^j_i=\diag{\exp(m^{\purple{j-1}}_i-m^{\purple{j}}_i)} O^{j-1}_i + \widetilde P^j_i V_j 
\end{align}

Where $m_i^j$ and $l_i^j$ are $b_q \times 1$ vectors, which are initialized to $- \infty$ and $0$ respectively. $\tilde \sigma()$ is an online softmax operator: $m^j_i = \max\{m^{j-1}_i, \rowmax(S^j_i)\},~\widetilde P_j^i=\exp(S^j_i-m^j_i)$.

Finally, \purple{after all iterations, i.e., $j = b_{kv}$}, the output $O_i$ can be computed by $O_i = \mathrm{diag}(l_i^j)^{-1} O_i^j$.

\subsection{Dynamic Quantization} \label{sec:dynamic_quant}

A matrix multiplication $C=AB$ can be accelerated with quantization as:
\begin{align}
(\delta_A, \hat A) = \psi(A),~~~~(\delta_B, \hat B) = \psi(B),~~~~\hat C=\hat A\hat B,~~~~C=\psi^{-1}_{\delta_A\delta_B}(\hat C)
\end{align}
Here, $\psi$ is a \emph{quantizer} which converts a high-precision (e.g., FP32) matrix $A$ to a low-precision format $\hat A$ (e.g., INT8 or FP8) with a \emph{scale} $\delta_A$, and $\psi^{-1}$ is a \emph{dequantizer} to convert back to high-precision. We should have $\psi^{-1}_{\delta_A}(\hat A)\approx A$. The actual matrix multiplication $\hat A\hat B$ is carried in low-precision. In modern GPUs, low-precision matrix multiplication is usually multiple times faster than higher-precision ones. 

Many quantizers depend on the numerical format and granularity, e.g., how many elements share a common scale factor. For example, an INT8 \emph{per-tensor dynamic quantizer} first computes the scale as the maximum absolute value of the entire tensor, scales the elements to the maximum representable range of INT8 [-127, +127], and then casts to INT8 with rounding: $\hat A = \lceil A/\delta_A \rfloor, \delta_A=\max(|A|)/127$. 
Likewise, \emph{per-token quantizer} assigns a scale factor for each token of a tensor: $\hat A[i,:] = \lceil A[i,:]/\delta_A \rfloor, \delta_A[i,:] = \max(|A[i,:]|)/127$. 
Also, \emph{per-channel quantizer} assigns a scale factor for each channel of the tensor, i.e., along the channel dimension: $A[:, i] = \lceil A[:, i] / \delta_A \rfloor, \delta_A = \max(|A[:, i]|) / {127}$.
Based on the tiling approach of FlashAttention, we can apply per-block quantization correspondingly. \emph{per-block quantizer} asigns a scale factor for every $b=m-n$ tokens: $\hat A[m:n,:] = \lceil A[m:n,:]/\delta_A \rfloor, \delta_A = \max(|A[m:n,:]|)/127$. 
Dequantization simply involves a element-wise scaling: $\psi_{\delta_A}^{-1}(\hat A)=\delta_A \hat A$.

\section{SAGE attention}

In this section, we propose \our, a fast yet accurate method to accelerate attention computation with 8-bit quantization. Considering that most networks are not natively trained with quantized attention, \our is designed to be plug-and-play. 

Unlike linear layers, which are easy to quantize, quantizing attention is more complicated. Extra treatment is required to ensure both good accuracy and fast speed. 
First, we will formulate quantized attention in Section~\ref{sec:formula}, followed by introducing our approach.

\subsection{Formulation} \label{sec:formula}
Based on the description of FlashAttention and dynamic quantization in Section~\ref{sec:flash_attn} and~\ref{sec:dynamic_quant}, we formulate the quantized attention as follows.

{
\small
\begin{align}
\label{equ:quant}
\textbf{Quantization: } \jt{(\delta_Q, \hat Q)} = \jt{\psi_Q}(Q/\sqrt{d}),~~\jt{(\delta_K, \hat K)} = \jt{\phi_K}(K),~~\jt{(\delta_P, \hat P)} = \jt{\psi_P}(\widetilde P),~~\jt{(\delta_V, \hat V)} = \jt{\psi_V}(V)\\
\label{equ:dequant}
\textbf{Attention: } ~~S = \jt{\psi^{-1}_{\delta_Q\delta_K}(\hat Q \hat K^\top)},~~(m^\prime, P) = \tilde\sigma(m, S),
~~O=\diag{\exp(m^\prime-m)} O + \jt{\psi^{-1}_{\delta_P\delta_V}(\hat P \hat V)}
\end{align}
}

$\phi_K$ is a transformation to obtain quantized $K$, which we shall discuss in subsequent sections. 
For simplicity, we omit all superscripts and subscripts, but the matrices used in attention are still tiles, and the computation is still organized as FlashAttention described in Section~\ref{sec:flash_attn}. Compared to the original full-precision version, as shown in Eq.~\ref{equ:quant},~\ref{equ:dequant}, \our adds quantizers to $Q,K,P,V$ and dequantizers to the product to accelerate both Matmuls of $QK^\top$ and $PV$. Online softmax is left in full-precision.

\begin{figure}[htb!]
    \begin{center}
    \includegraphics[width=1.0\textwidth]{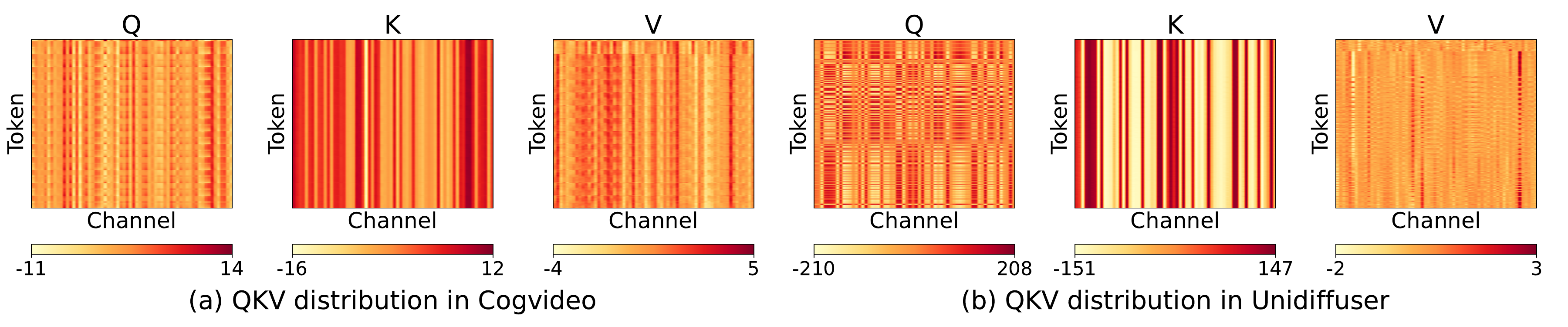}
    \vspace{-1em}
    \caption{Typical examples of data distribution of (Q, K, V).}
    \end{center}
    \label{fig:heatmap}
\end{figure}

\begin{table}[htb!]
    \caption{End-to-end metrics comparison of different quantization methods.}
    \label{tab:quant_type_analysis}
    \setlength\tabcolsep{2.8pt}
    \begin{center}
    \scalebox{0.97}{
    \begin{tabular}{c|c|c|c|c|c|c}
    \toprule
    \bf\makecell{Quantization \\ (Q, K)} & \bf\makecell{Smoothing \\ K} & \bf\makecell{Llama \\ WikiText} $\downarrow$ & \bf\makecell{CogVideo \\ (Fscore)} $\uparrow$ & \bf\makecell{Unidiffuser \\ (FID)} $\downarrow$ & \bf\makecell{UltraPixel \\ (FID)} $\downarrow$ & \bf\makecell{TIMM \\ ImageNet} $\uparrow$ \\
    \hline
    \bf{Full-Precision} &         -             & 5.823 & 3.768 & 163.33 & 179.78 & 84.79\%    \\ \hline
    \bf\multirow{2}{*}{Per-token}  & \ding{55}  & 5.824 & \dred{1.924} & \dred{221.18} & \dred{193.36} & \dred{84.21\%} \\
                                & \cellcolor{gray!24}\checkmark & \cellcolor{gray!24}5.824 & \cellcolor{gray!24}\textbf{\dgreen{3.734}} & \cellcolor{gray!24}\textbf{\dgreen{166.52}} & \cellcolor{gray!24}\textbf{\dgreen{179.79}} & \cellcolor{gray!24}\dgreen{84.74\%} \\ \hline
    \bf\multirow{2}{*}{Per-block}  & \ding{55}  & 5.825 & \dred{2.014} & \dred{229.08} & \dred{195.67} & \dred{84.18\%} \\
                                & \cellcolor{gray!24}\checkmark & \cellcolor{gray!24}5.824 & \cellcolor{gray!24}\textbf{\dgreen{3.718}} & \cellcolor{gray!24}\textbf{\dgreen{166.93}} & \cellcolor{gray!24}\textbf{\dgreen{179.98}} & \cellcolor{gray!24}\dgreen{84.76\%} \\ \hline
    \bf\multirow{2}{*}{Per-tensor} & \ding{55}  & 5.826 & \dred{1.902} & \dred{267.06} & \dred{196.26} & \dred{84.12\%} \\
                                & \cellcolor{gray!24}\checkmark & \cellcolor{gray!24}5.824 & \cellcolor{gray!24}\textbf{\dgreen{3.640}} & \cellcolor{gray!24}\textbf{\dgreen{167.65}} & \cellcolor{gray!24}\textbf{\dgreen{180.21}} & \cellcolor{gray!24}\dgreen{84.69\%} \\ \hline
    \multicolumn{2}{c|}{\bf FlashAttn3 (with quant)}  & \dred{5.850} &  \dred{3.394}  & \dred{394.13} & \dred{383.61} &  84.70\% \\
    \bottomrule
    \end{tabular}
    }
    \end{center}
\end{table}

\subsection{Smooth Matrix K} \label{sec:smooth_k}
 
Directly quantizing $Q, K$  often results in a large error. Particularly, quantizing $Q, K$ to INT8 yields completely blurry image/video in text-to-image/video tasks. 
As shown in Figure~\ref{fig:heatmap}, we visualize two typical groups of $Q, K, V$ from a text-to-image model \unidiffuser~\citep{bao2022uni} and a text-to-video model \cogvideo~\citep{yang2024cogvideox}. Notably, $K$ exhibits distinct channel-wised outliers. However, per-channel quantization cannot be applied for $K$, because quantization can only be performed at the outer axis (token dim) of the Matmul $QK^\top$. Moreover, the previous smoothing technique proposed for linear layers ~\citep{xiao2023smoothquant} cannot be applied since $Q$ is also heavily affected by outliers. Fortunately, the channel outliers of $K$ have a pattern: Each token's key is actually a \emph{large bias shared by all tokens}, plus a small token-wise signal. Therefore, the outlier is not from large variation across tokens, but simply the large bias. 
Based on this observation, we propose to smooth the matrix $K$ by a transform $\gamma$, which subtracts averaged $K$ across all tokens:
\begin{align} 
    \gamma(K) = K - \mean(K)
\end{align}
where $\mean(K) = \frac{1}{N}\sum_{t=1}^N K[t,:]$ is the average key, with a shape $1\times d$.
Note that such a transformation does not change the attention score $P$, because for any query $q$, we have $\sigma(q (K-\mean(K)^\top)) = \sigma(q K^\top - q\cdot \mean(K)) = \sigma(qK^\top)$. Finally, the transformation from full-precision $K$ to quantized $\hat K$ can be written as $\phi_K(K) = \psi_K\circ \gamma$, where $\psi_K$ is a quantizer. In other words, a full-precision $K$ is substracted with the mean, before eventually being quantized.

Table~\ref{tab:quant_type_analysis} presents end-to-end metrics for different quantization methods with and without \textit{smoothing K} on various models. The results demonstrate that \textit{smoothing K} offers significant benefits of accuracy. Moreover, the speed overhead of smoothing $K$ for attention is less than \textbf{0.2\%} (See Table\ref{exp:overhead_km}).

\begin{table}[htb!]
    \centering
    \begin{minipage}{0.49\textwidth}
        \raggedright
        \caption{\textbf{Average accuracy} using different data types across all layers of real models.}
        \label{tab:quant_precision_analysis1}
        \setlength\tabcolsep{2.1pt}
        \scalebox{0.91}{
        \begin{tabular}{c|c|c|c|c}
            \toprule
            $Q, K$ & $\widetilde P, V$ & \textbf{Cos Sim $\uparrow$} & \textbf{Relative L1 $\downarrow$} & \textbf{RMSE $\downarrow$} \\
            \hline
            & E4M3 & 99.94\% & 0.0345 & 3.53e-3 \\
            \cellcolor{gray!24}\textbf{INT8} & E5M2 & 99.81\% & 0.0572 & 6.11e-3 \\
                                              & INT8 & 99.70\% & 0.1035 & 6.82e-3 \\
            \hline
            \multirow{3}{*}{E4M3} & E4M3 & 99.81\% & 0.0607 & 5.93e-3 \\
                                              & E5M2 & 99.68\% & 0.0769 & 7.72e-3 \\
                                              & INT8 & 99.58\% & 0.1199 & 8.31e-3 \\
            \hline
            \multirow{3}{*}{E5M2} & E4M3 & 99.37\% & 0.1107 & 1.09e-2 \\
                                              & E5M2 & 99.22\% & 0.1213 & 1.20e-2 \\
                                              & INT8 & 99.13\% & 0.1583 & 1.24e-2 \\
            \bottomrule
        \end{tabular}
        }
    \end{minipage}
    \hspace{0.01\textwidth} 
    \begin{minipage}{0.46\textwidth}
        \raggedleft
        \caption{\textbf{Worst accuracy} using different data types across all layers of real models.}
        \label{tab:quant_precision_analysis2}
        \setlength\tabcolsep{2.1pt}
        \scalebox{0.91}{
        \begin{tabular}{c|c|c|c|c}
            \toprule
            $Q, K$ & $\widetilde P, V$ & \textbf{Cos Sim $\uparrow$} & \textbf{Relative L1 $\downarrow$} & \textbf{RMSE $\downarrow$} \\
            \hline
            \multirow{4}{*}{INT8}  & E4M3 & 76.36\%	   & 0.5899	   & 0.4311 \\
                                   & E5M2 & 78.98\%	& 0.4233	& 0.4371 \\
                                   & INT8 & 56.40\%	& 0.7921	& 0.5405 \\ \cline{2-5}
                                   & \cellcolor{gray!21}\bf{FP16} & \cellcolor{gray!24}\dgreen{99.99\%}	& \cellcolor{gray!24}\dgreen{0.0116}	& \cellcolor{gray!24}\dgreen{0.0091} \\
            \bottomrule
                                       
        \end{tabular}
        }
    \end{minipage}
\end{table}

\subsection{Quantization for Q, K, P, V} \label{sec:qkpv_quant}

\textbf{Quantization granularity for $Q, K$}: 
$\psi_Q(Q)$ and $\psi_K(K)$ can be set with the granularity of per-token, per-block or per-tensor. This is because per-channel quantization is not feasible, since the scale factors of the inner axis of $Q K^\top$ cannot be used to do dequantization~\citep{xiao2023smoothquant}.

\textbf{Data type of $Q, K$}: We choose INT8 for $\psi_Q(Q)$ and $\psi_K(K)$ for two reasons. First, Table~\ref{tab:quant_precision_analysis1} shows the average accuracy using different data types (INT8, E4M3, E5M2) for $Q, K, \widetilde P, V$ across all layers of \llama(7B)~\citep{touvron2023llama2} and \unidiffuser. It shows that quantizing $Q, K$ to INT8 performs higher accuracy than using E4M3 and E5M2.
Second, Matmul using INT8 is two times faster than using FP8 in many commonly used GPUs, e.g., RTX4090 and 3090.

\textbf{Quantization granularity for $\widetilde P, V$}: We propose to use $\psi_P(\widetilde P)$ in per-block and $\psi_V(V)$ in per-channel for three reasons. (1) Per-channel quantization for $\widetilde P$ and per-token quantization for $V$ are not viable because dequantization requires scale factors of the outer axis. (2) $\widetilde P = \mathrm{exp}(S_i - \mathrm{rowmax}(S_i))$, where $S_i$ is the Matmul result of a block of $Q$ and $K^T$, the max value in each row of $\widetilde P$ is 1. Hence, we can assign a single static scale $s = \frac{1}{127}$ to a block $\widetilde P$, whose accuracy equals per-token quantization. (3) Per-channel quantization can address the channel-wised outlier of $V$.

\textbf{Data type of $\widetilde P, V$}: We choose INT8 for $\psi_P(\widetilde P)$ and $\psi_V(V)$ because Matmul using INT8 is two times faster than using FP8 in some commonly used GPUs, and although the accuracy using $\psi_P(\widetilde P)$ and $\psi_V(V)$ in INT8 is worse than E4M3 and E5M2, the average accuracy is similar (See Table~\ref{tab:quant_precision_analysis1}).

\textbf{Accuracy metrics.} We use three metrics to assess the accuracy of quantized attention output $O'$ compared to attention output in full-precision $O$: First, we flatten $O'$ and $O$ into vectors in the shape of $1\times n$. Then, Cosine Sim$=\sum OO' / \smash{\sqrt{\sum O^2}} \smash{\sqrt{\sum O'^2}}$, Relative L1$=\sum |O - O'| / \sum |O|$, RMSE$=\sqrt{(1/n) \sum (O - O')^2}$.

\begin{table}[htb!]
    \centering
    \begin{minipage}{0.46\textwidth}
        \raggedright
        \caption{\textbf{Average accuracy} using different accumulators across all layers of real models.}
        \label{tab:quant_accu_analysis1}
        \setlength\tabcolsep{5pt}
        \scalebox{0.93}{
        \begin{tabular}{c|c|c|c}
            \toprule
            \textbf{Accum.} & \textbf{Cos Sim $\uparrow$} & \textbf{Relative L1 $\downarrow$} & \textbf{RMSE $\downarrow$} \\
            \hline
            \textbf{FP32} & 99.98\% & 0.0156 & 2.94e-3 \\
            \cellcolor{gray!24}\textbf{FP16} & \cellcolor{gray!24}99.98\% & \cellcolor{gray!24}0.0156 & \cellcolor{gray!24}2.94e-3 \\
            \bottomrule
        \end{tabular}
        }
    \end{minipage}
    \hspace{0.025\textwidth} 
    \begin{minipage}{0.5\textwidth}
        \raggedleft
        \caption{\textbf{Worst accuracy} using different accumulators across all layers of real models.}
        \label{tab:quant_accu_analysis2}
        \setlength\tabcolsep{5pt}
        \scalebox{0.93}{
        \begin{tabular}{c|c|c|c}
            \toprule
            \textbf{Accum.} & \textbf{Cos Sim $\uparrow$} & \textbf{Relative L1 $\downarrow$} & \textbf{RMSE $\downarrow$} \\
            \hline
            \textbf{FP32} & 99.84\% & 0.0511 & 4.229e-3 \\
            \cellcolor{gray!24}\textbf{FP16} & \cellcolor{gray!24}99.84\% & \cellcolor{gray!24}0.0511 & \cellcolor{gray!24}4.229e-3 \\
            \bottomrule
        \end{tabular}
        }
    \end{minipage}
\end{table}

\subsection{FP16 accumulator: much more accurate and efficient solution} \label{sec:fp16_acc}

The above solution for $\psi_P(\widetilde P)$ and $\psi_V(V)$ has one problem, that is, the accuracy using INT8 is very poor in some model layers. Table~\ref{tab:quant_precision_analysis2} shows the worst accuracy using different data types for $Q, K, \widetilde P, V$ across all layers
of \llama and \unidiffuser. It shows that INT8 $\psi_P(\widetilde P)$ and $\psi_V(V)$ bring an unacceptable error. 
In response, we propose a very accurate and also efficient solution. Specifically, we propose to use FP16 as the data type of Matmul $\widetilde P V$ with an FP16 accumulator. 

The benefit of such a solution is obvious. First, in the context of some commonly used GPUs, e.g., RTX4090 and 3090, the speed of Matmul in FP16 with an FP16 accumulator is \textbf{2x} faster than that with an FP32 accumulator. Moreover, using FP16 accumulators can save more register resources than using FP32 accumulators, accelerating the computation speed. 
Second, Table~\ref{tab:quant_precision_analysis2} shows that using FP16 for $\widetilde P, V$ is much more accurate than using all the other 8-bit data types. Moreover, using FP16 accumulators incurs no accuracy loss than using FP32 accumulators. Specifically, Table~\ref{tab:quant_accu_analysis1} and~\ref{tab:quant_accu_analysis2} show the average and worst accuracy using FP16 or FP32 accumulators on all layers of \llama and \unidiffuser, showing that there is no accuracy loss of using the FP16 accumulator.

\begin{table}[htb!]
    \centering
    \vspace{-.5em}
    \caption{Four kernel implementations of \our.}
    \label{tab:adaptive_types}
    \setlength\tabcolsep{5pt}
    \scalebox{0.91}{
    \begin{tabular}{c|c|c|c}
        \toprule
        \textbf{Kernel} & $\psi_Q(Q)$, $\psi_K(K)$ & $\psi_P(P)$ & $\psi_V(V)$ \\
        \hline
        \ourt  & per-token, INT8	 & FP16, FP16 Accumulator  & FP16, FP16 Accumulator \\
        \ourb (Algorithm~\ref{alg:ourb}) & per-block, INT8	 & FP16, FP16 Accumulator  & FP16, FP16 Accumulator\\
        \ourvt (Figure~\ref{fig:method_overview}(a)) & per-token, INT8	 & per-block, INT8	       & per-channel, INT8	 \\ 
        \ourvb & per-block, INT8     & per-block, INT8	       & per-channel, INT8 \\
        \bottomrule                  
    \end{tabular}
    }
\end{table}

\begin{figure}[h]
    \begin{center}
    \includegraphics[width=0.95\textwidth]{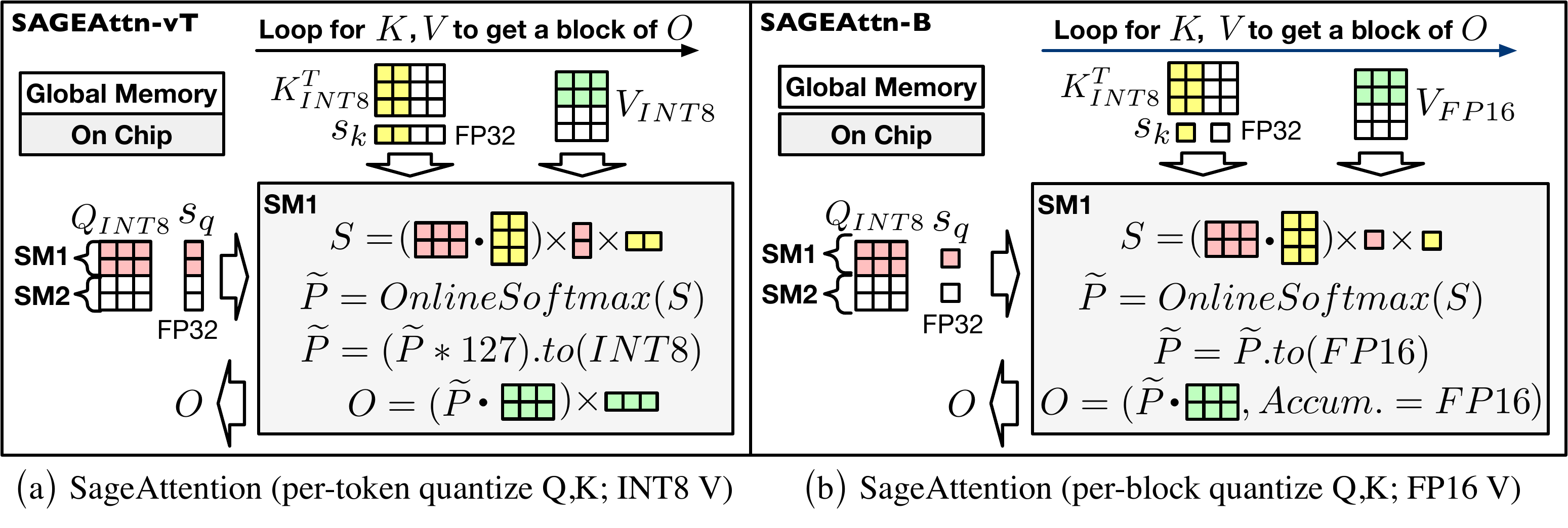}
    \vspace{-.9em}
    \caption{Workflow of SageAttention.}
    \label{fig:method_overview}
    \end{center}
\end{figure}

\setlength{\textfloatsep}{5pt}
\begin{algorithm}[htb]  
    \small
    \caption{Implementation of \ourb.}
    \label{alg:ourb} 

    \KwIn{Matrices $Q(\text{FP16}), K(\text{FP16}), V(\text{FP16}) \in \mathbb{R}^{N \times d}$, block size $b_q, b_{kv}$.}

    \textbf{Preprocessing:} \jt{$K=K-\mathrm{mean}(K)$} ; \tcp{\annotate{Subtracting the mean value across tokens}}

    \textbf{Quantization:} $\jt{(\delta_Q, \hat Q)} = \jt{\psi_Q}(Q/\sqrt{d}),~~\jt{(\delta_K, \hat K)} = \jt{\psi_K}(K)$ ; \tcp{\annotate{INT8 per-block quant}}

    Divide \jt{$\hat Q$} into $T_m = {N}/{b_q}$ blocks \jt{$\{\hat Q_i\}$}, and divide \jt{$\hat K$}, $V$ into $T_n = {N}/{b_{kv}}$ blocks \jt{$\{\hat K_i\}$} and $\{V_i\}$;  

    \For (\tcp*{\annotate{Outer loop is paralleled in SMs (stream processors)}}) {$\textbf{i}$ in [1, $T_m$]}{ \vspace{-1em}
        Load \jt{$\hat Q_i$} and \jt{$\delta_Q[i]$} into a SM ;

        \For {$\textbf{j}$ in [1, $T_n$]} { 
            Load \jt{$\hat K_j$}, $V_j$, and \jt{$\delta_K[j]$} into the SM ;

            \jt{$S_i^j = \mathrm{Matmul}(\hat Q_i, \hat K_j^T) \times \delta_Q[i] \times \delta_K[j]$;}

            $m_i^j = \mathrm{max}(m_i^{j-1}, \mathrm{rowmax}(S_i^j))$, $ \widetilde P_i^j = \mathrm{exp}(S_i^j - m_i^j)$, $l_i^j = e^{m_i^{j-1}-m_i^j} l_i^{j-1} + \mathrm{rowsum}(\widetilde P_i^j)$ ;

            $O_i^j = \mathrm{diag}(e^{m_i^{j-1}-m_i^j}) O_i^{j-1} +$ \jt{$\mathrm{Matmul}(\widetilde P_i^j\text{.to(FP16)}, V_j, \text{Accum\_type = FP16})$} ;
        }

        $O_i = \mathrm{diag}(l_i^{T_n})^{-1} O_i^{T_n}$ ;

        Write $O_i$ ;  
    }
    \textbf{return} $O = \{O_i\}$;
\end{algorithm}

\subsection{Adaptive Quantization} \label{sec:adaptive}

Based on the discussion in Section~\ref{sec:qkpv_quant} and ~\ref{sec:fp16_acc}, we implement four attention kernels (See Table~\ref{tab:adaptive_types}) based on two sets of choices: (1) Using $\psi_Q(Q)$ and $\psi_K(K)$ in per-token or per-block. (2) Using $\psi_P(\widetilde P)$ and $\psi_V(V)$ in INT8 or retaining $\widetilde P, V$ in FP16 with an FP16 accumulator.
\purple{\ourb is accurate enough for all models and can achieve a 2x speedup (See Figure~\ref{fig:kf_h64_baseline} and~\ref{fig:kf_h128_baseline}). However, \ourvb is also accurate for some layers in a model and faster a little (about 4\%) than \ourb. Therefore, we use various inputs to test the cosine similarity of \ourvb for each layer of a model. Then, we will select \ourvb for those layers where \ourvb's cosine similarity is bigger than 99.8\% (the worst similarity of \ourb), and the other layers are left for \ourb.}

\subsection{Fusion Tricks and Performance Analysis} \label{sec:fuse_and_performance}

\textbf{Fusion Tricks.} To reduce the overhead of quantization, we fuse the quantization process with the operator preceding the attention layer. For instance, we fuse quantization within the ROPE (Rotary Position Embedding)~\citep{su2021roformer} layer. Specifically, before the ROPE result ($A$) is written from shared memory into global memory, we perform $\delta_A, \hat A = \psi(A)$. Subsequently, the $\delta_A, \hat A$ are written into global memory. Additionally, we also fuse the coefficient ($1/{\sqrt{d}}$) of $Q K^T$ into the quantization process rather than leaving it in the attention layer. Specifically, we multiply $Q$ by ($1/{\sqrt{d}}$) on chip before quantizating $Q$.
\textbf{Performance Analysis.} We will take \ourb as an example to discuss the acceleration effects on actual hardware:
(1) \underline{Matmul acceleration.} Utilizing INT8 matrix multiplication units on current mainstream hardware can achieve \textbf{2-4}$\times$ throughput. While FP16 accumulators do not offer throughput improvements on most compute cards, on-edge accelerators, such as the RTX4090, can still achieve a 2x improvement over FP32 accumulators.
(2) \underline{Quantization overhead.}
Quantization and dequantization are considered the main overhead in current quantization methods~\citep{lin2024qservew4a8kv4quantizationcodesign}. The computational overhead can not be avoided, but through fusing the quantization of $Q, K$ with ROPE, we avoid the IO overhead of quantization. 
(3) \underline{Cache and registers.} Currently, mainstream accelerators need to store data in a cache (such as SharedMemory) during computation. Using 8-bit data for calculations can reduce the usage of the general cache, and using fp16 accumulators can also reduce the usage of accumulation registers. 
(4) \underline{Dram access.} Using 8-bit data can halve the tensors transfer overhead from DRAM to the compute units. Although quantization introduces additional FP32 scales, these scales can be considered negligible compared to the tensors.

\section{Experiments}

\noindent \textbf{Main results.} The speed of \our is approximately $\textbf{2.1}\times$ faster than FlashAttention-2. Furthermore, \our achieves an average real speedup of $\textbf{2.83}\times$ compared to the original attention in various models, with \textbf{negligible loss in end-to-end metrics}.

\subsection{Experimental Setup} \label{sec:exp_set_up}

\noindent \textbf{Models.} 
We validate the effectiveness of \our across a diverse set of representative models from the fields of language, image, and video generation.
Specifically, we conduct experiments on five models: \llama(7B)~\citep{touvron2023llama2} for text2text, \cogvideo~\citep{yang2024cogvideox} for text2video, \unidiffuser~\citep{bao2022uni} and \ultrapixel~\citep{ren2024ultrapixel} for text2image, \timm~\citep{rw2019timm} for image classification, \purple{and \llava~\citep{liu2024llavanext} for visual question answering.}

\noindent \textbf{Datasets.}   
\llama is evaluated on three zero-shot tasks: WikiText~\citep{merity2022pointer} to assess the model's prediction confidence, LAMBADA~\citep{paperno2016lambada} evaluate contextual understanding, and MMLU~\citep{2020mmlu} for measuring knowledge across various subjects.
\cogvideo is evaluated using the open-sora~\citep{opensora} prompt sets.
Both \ultrapixel and \unidiffuser are assessed on the COCO annotations~\citep{lin2014microsoft}, featuring (prompt, image) pairs.
\timm is evaluated on on three image datasets: ImageNet~\citep{deng2009imagenet}, ImageNet-Sketch (Sketch)~\citep{wang2019sketch}, and ImageNet-Rendition (ImageNet-r)~\citep{hendrycks2021rendition}. \purple{\llava is evaluated on three datasets: TextVQA~\citep{singh2019towards}, POPE~\citep{li2023evaluating}, and VQAv2~\citep{goyal2017making}.}

\noindent \textbf{Metrics.}   
For \llama, we use perplexity (ppl.)~\citep{jelinek1977perplexity} for WikiText, and Accuracy (Acc.) for LAMBADA and MMLU.
For \cogvideo, folowing~\citep{zhao2024vidit}, we evaluate the quality of generated videos on five metrics: CLIPSIM and CLIP-Temp (CLIP-T)~\citep{liu2024evalcrafter} to measure the text-video alignment; (VQA-a) and (VQA-t) to assess the video aesthetic and technical quality, respectively; and Flow-score (FScore) for temporal consistency~\citep{wu2023exploring}. 
For \ultrapixel and \unidiffuser, generated images are compared with the images in the COCO annotations dataset in three aspects: FID~\citep{heusel2017gans} and sFID~\citep{salimans2016improved} for fidelity evaluation, \textit{Clipscore} (CLIP)~\citep{hessel2021clipscore} for text-image alignment, and \textit{ImageReward} (IR)~\citep{xu2024imagereward} for human preference.
For \timm and \llava, we use Accuracy.

\begin{figure}[htb!]
    \centering
    \vspace{-.2em}
    \includegraphics[width=1.0\textwidth]{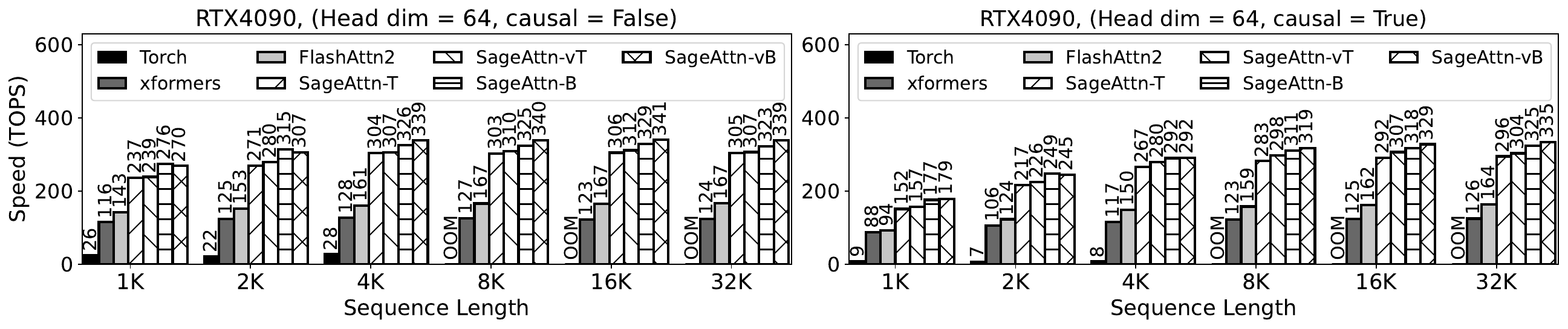}
    \vspace{-1.8em}
    \caption{Speed comparison between \our and baselines (RTX4090, headdim=64).}
    \vspace{-.5em}
    \label{fig:kf_h64_baseline}
\end{figure}

\begin{figure}[htb!]
    \centering
    \vspace{-.2em}
    \includegraphics[width=1.0\textwidth]{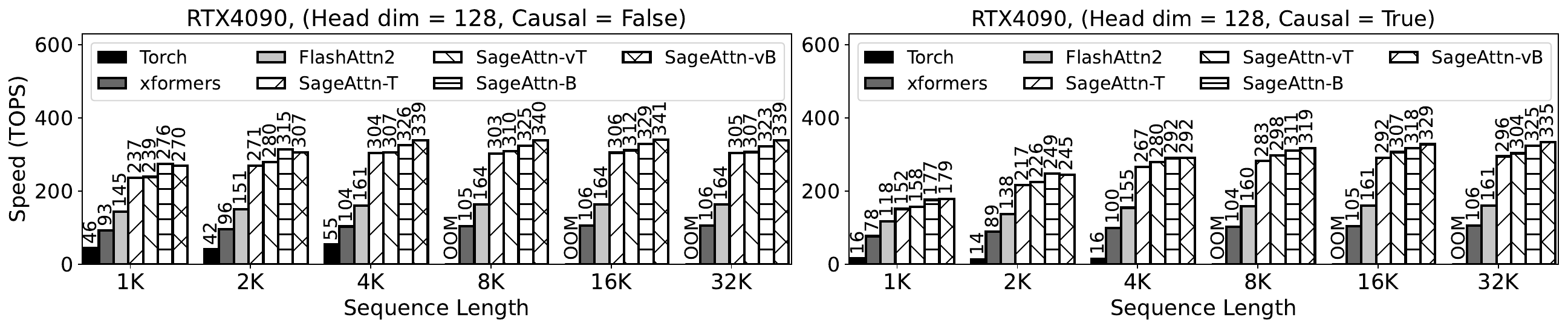}
    \vspace{-1.8em}
    \caption{Speed comparison between \our and baselines (RTX4090, headdim=128).}
    \vspace{-.5em}
    \label{fig:kf_h128_baseline}
\end{figure}

\begin{figure}[htb!]
    \centering
    \vspace{-.2em}
    \includegraphics[width=1.0\textwidth]{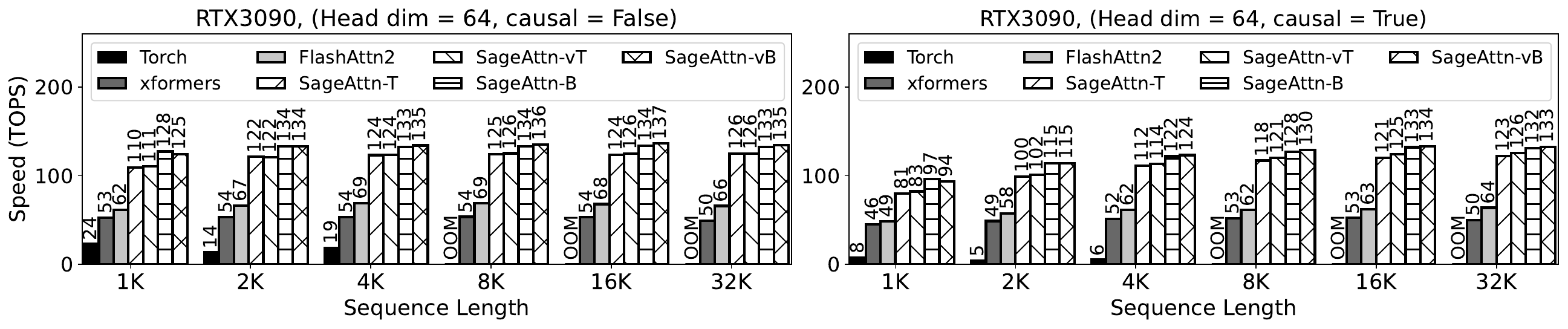}
    \vspace{-1.8em}
    \caption{Speed comparison between \our and baselines (RTX3090, headdim=64).}
    \vspace{-.5em}
    \label{fig:kf_h64_baseline_3090}
\end{figure}

\begin{figure}[htb!]
    \centering
    \vspace{-.2em}
    \includegraphics[width=1.0\textwidth]{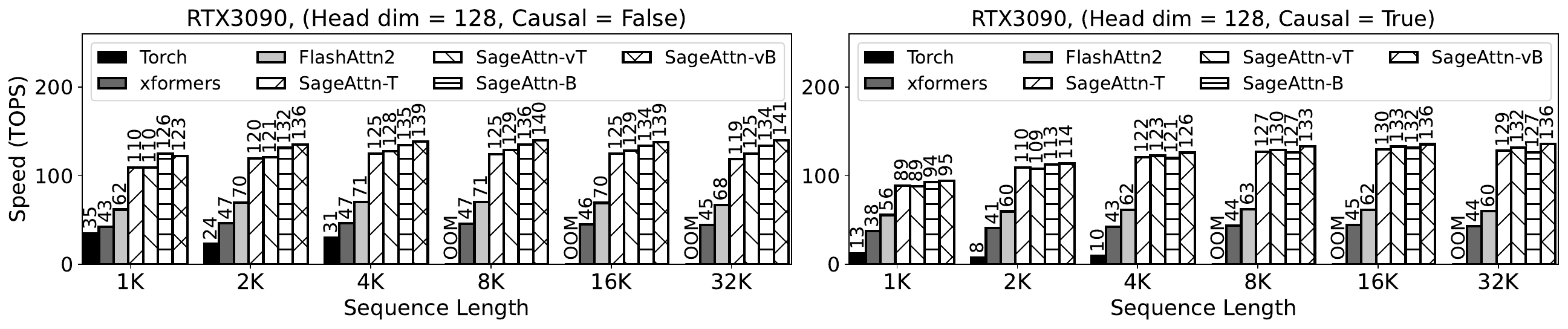}
    \vspace{-1.8em}
    \caption{Speed comparison between \our and baselines (RTX3090, headdim=128).}
    \label{fig:kf_h128_baseline_3090}
\vspace{.2em}
\end{figure}

\subsection{Speed and Accuracy of attention Kernels}

\noindent \textbf{Speed.} We conduct experiments to compare the Speed of \our against baselines using configurations with headdim=64 or headdim=128, both with and without Causal Mask~\cite{vaswani2017attention}. Specifically, Figure~\ref{fig:kf_h64_baseline} and Figure~\ref{fig:kf_h128_baseline} show the Speed of \our and baselines across varying sequence lengths on RTX4090. These results indicate that \our achieves a peak of \textbf{341} TOPS and is \textbf{2x} faster than FlashAttention2 and \textbf{2.9x} faster than xformers on average.
Figure~\ref{fig:kf_h64_baseline_3090} and Figure~\ref{fig:kf_h128_baseline_3090} illustrate the results on RTX3090, showing a similar speedup performance.

\noindent \textbf{Accuracy.} 
Table~\ref{tab:error of sage} shows the numerical error of four implementations of \our compared with attention in full-precision. This experiment is conducted using a set of (Q, K, V) conforming to a normal distribution. It shows the error of the four implementations is rather small. \ourt and \ourb achieve 100\% cosine similarity and RMSE in the e-4 level.

\begin{table}[htb!]
\vspace{-.2em}
    \caption{Real speedup of \our on RTX4090.}
    \label{exp:real_flops}
    \setlength\tabcolsep{6.5pt}
    \begin{center}
    \begin{tabular}{l|c|l|c|c}
    \toprule
    {\bf Model}  & {\bf Shape of Q, K, V}  & {\bf Original attention}  & {\bf SageAttention}  & {\bf Speedup}
    \\ \hline
    \cogvideo    & (2, 30, 17776, 64) &  163.37 (FlashAttn2)  &  \cellcolor{gray!20}\textbf{327.57}  & \cellcolor{gray!21}\textbf{2.01x}  \\
    \llama       & (4, 32, 1536, 128) &  130.99 (FlashAttn2)  &  \cellcolor{gray!20}\textbf{231.74}  & \cellcolor{gray!21}\textbf{1.77x}  \\ 
    \ultrapixel  & (2, 32, 7285, 64)  &  152.03 (FlashAttn2)  &  \cellcolor{gray!20}\textbf{325.18}  & \cellcolor{gray!21}\textbf{2.14x}  \\ 
    \unidiffuser & (4, 24, 1105, 64)  &  105.68 (xformers)    &  \cellcolor{gray!20}\textbf{246.93}  & \cellcolor{gray!21}\textbf{2.34x}  \\ 
    \timm        & (12, 64, 197, 64)  &  18.910 (Torch)       &  \cellcolor{gray!20}\textbf{111.41}  & \cellcolor{gray!21}\textbf{5.89x}  \\ \bottomrule
    \end{tabular}
    \end{center}
\vspace{-.2em}
\end{table}

\begin{table}[htb!]
    \caption{End-to-end metrics loss across text, image, and video generation models.}
    \label{exp:metrics_loss_t2t}
    \setlength\tabcolsep{4.25pt}
    \begin{center}
    \begin{tabular}{p{2.1cm}|p{2.55cm}|c|c|c}
    \toprule
    {\bf Model}  & {\bf Attention}  & {\bf WikiText (Ppl.) $\downarrow$}  & {\bf Lambda (Acc.) $\uparrow$}  & {\bf MMLU (Acc.) $\uparrow$}  \\ \hline
    \multirow{2}{*}{\llama} & Full-Precision & 5.823 & 0.886  & 0.46   \\  
    & \cellcolor{gray!24}\textbf{SageAttention} & \cellcolor{gray!24}\textbf{\dgreen{5.824}} &  \cellcolor{gray!24}\textbf{\dgreen{0.887}}  & \cellcolor{gray!24}\textbf{\dgreen{0.46}}   \\ \bottomrule
    \end{tabular}
    \end{center}
    
    \begin{center}
    \setlength\tabcolsep{3.6pt}
    \begin{tabular}{p{2.2cm}|p{2.55cm}|c|c|c|c|c}
    \toprule
    {\bf Model}  & {\bf Attention}  & {\bf CLIPSIM $\uparrow$}  & {\bf CLIP-T $\uparrow$}  & {\bf VQA-a $\uparrow$}  & {\bf VQA-t $\uparrow$}  & {\bf FScore $\uparrow$} \\ \hline
    \multirow{2}{*}{\cogvideo} & Full-Precision & 0.1837 & 0.9976  & 68.962  & 75.925  &  3.7684  \\  
    & \cellcolor{gray!24}\textbf{SageAttention} & \cellcolor{gray!24}\textbf{\dgreen{0.1836}} &  \cellcolor{gray!24}\textbf{\dgreen{0.9976}}  & \cellcolor{gray!24}\textbf{\dgreen{68.839
    }}  &  \cellcolor{gray!24}\textbf{\dgreen{75.037}}  &  \cellcolor{gray!24}\textbf{\dgreen{3.8339}} \\ \bottomrule
    \end{tabular}
    \end{center}
    
    \begin{center}
    \setlength\tabcolsep{15.5pt}
    \begin{tabular}{p{1.4cm}|p{1.725cm}|c|c|c|c}
    \toprule
    {\bf Model}  & {\bf Attention}  & {\bf FID $\downarrow$}  & {\bf sFID $\downarrow$}  & {\bf CLIP $\uparrow$}  & {\bf IR $\uparrow$}
    \\ \hline
    \multirow{2}{*}{\hspace{-1.2em}\unidiffuser} & \hspace{-1.2em}Full-Precision & 163.33 & 145.08  & 0.3152  & 0.1609  \\  
    & \cellcolor{gray!24}\hspace{-1.2em}\textbf{SageAttention} &  \cellcolor{gray!24}\textbf{\dgreen{166.49}} & \cellcolor{gray!24}\textbf{\dgreen{143.18}}  & \cellcolor{gray!24}\textbf{\dgreen{0.3154}}  &  \cellcolor{gray!24}\textbf{\dgreen{0.1521}} \\ \hline
    \multirow{2}{*}{\hspace{-1.2em}\ultrapixel} & \hspace{-1.2em}Full-Precision & 179.78 & 141.35  & 0.3132  & 0.6169  \\  
    & \cellcolor{gray!24}\hspace{-1.2em}\textbf{SageAttention} & \cellcolor{gray!24}\textbf{\dgreen{179.79}} &  \cellcolor{gray!24}\textbf{\dgreen{141.63}} & \cellcolor{gray!24}\textbf{\dgreen{0.3131}}  &  \cellcolor{gray!24}\textbf{\dgreen{0.6110}} \\ \bottomrule
    \end{tabular}
    \end{center}

    \begin{center}
    \setlength\tabcolsep{2.3pt}
    \begin{tabular}{p{2.35cm}|p{2.65cm}|c|c|c}
    \toprule
    {\bf Model}  & {\bf Attention}  & {\bf ImageNet (Acc.) $\uparrow$}  & {\bf Sketch (Acc.) $\uparrow$}  & {\bf ImageNet-r (Acc.) $\uparrow$}  \\ \hline
    \multirow{2}{*}{\timm} & Full-Precision &  84.79\% & 45.32\%  & 59.55\%   \\  
            & \cellcolor{gray!24}\textbf{SageAttention} & \cellcolor{gray!24}\textbf{\dgreen{84.74\%}}  &  \cellcolor{gray!24}\textbf{\dgreen{45.78\%}}  & \cellcolor{gray!24}\textbf{\dgreen{60.32\%}}   \\ \bottomrule
    \end{tabular}
    \end{center}

    \begin{center}
    \setlength\tabcolsep{5.65pt}
    \begin{tabular}{p{2.15cm}|p{2.4cm}|c|c|c}
    \toprule
    {\bf Model}  & {\bf Attention}  & {\bf TextVQA (Acc.) $\uparrow$}  & {\bf POPE (Acc.) $\uparrow$}  & {\bf VQAv2 (Acc.) $\uparrow$}  \\ \hline
    \multirow{2}{*}{\llava} & Full-Precision & 60.25\% & 86.45\%  & 77.55\%   \\  
            & \cellcolor{gray!24}\textbf{SageAttention} & \cellcolor{gray!24}\textbf{\dgreen{60.09\%}}  &  \cellcolor{gray!24}\textbf{\dgreen{86.44\%}}  & \cellcolor{gray!24}\textbf{\dgreen{77.47\%}}   \\ \bottomrule
    \end{tabular}
    \end{center}
\end{table}
\vspace{-1em}

\subsection{End-to-end Performance}
\textbf{Speedup.} We measure the real speed of \our and the original attention on \unidiffuser, \ultrapixel, \cogvideo, \llama and \timm on RTX4090. 
Table~\ref{exp:real_flops} shows that \our outperforms original attention across all models. Specifically, \our yields \textbf{2.83x} speedup compared to the original attentions on average.

\textbf{Metrics loss.} We assessed the end-to-end metrics of various models using \our compared to using attention in full-precision.
Detailed evaluation results are presented in Table~\ref{exp:metrics_loss_t2t} for \llama, \cogvideo, \unidiffuser, \ultrapixel, and \timm, respectively. The results indicate that \our successfully matches the performance of attention in full-precision across all models. Specifically, on \llama, \cogvideo, \ultrapixel, and \unidiffuser, \our resulted in only a minor average degradation of 0.2\% compared to attention in full-precision. Moreover, on \timm, \our even surpasses attention in full-precision.

\begin{table}[htb!]
\vspace{-1.3em}
    \centering
    \begin{minipage}{0.54\textwidth}
        \raggedright
        \caption{Accuracy of SageAttention kernels.}
        \label{tab:error of sage}
        \setlength\tabcolsep{2.7pt}
        \scalebox{0.95}{
        \begin{tabular}{c||c|c|c}
            \toprule
            {\bf attention}  & {\bf Cos Sim $\uparrow$}  & {\bf Relative L1 $\downarrow$}  & {\bf RMSE $\downarrow$}  \\ \hline
            \ourt   &   1.0    & 0.019  & 6.8e{-4}  \\ 
            \cellcolor{gray!24}\ourb   &   \cellcolor{gray!24}1.0   & \cellcolor{gray!24}0.021  & \cellcolor{gray!24}7.3e{-4}  \\ 
            \ourvt  &  99.9\%   & 0.064  & 0.065  \\ 
            \ourvb  &  98.9\%   & 0.138  & 0.067  \\ \bottomrule
        \end{tabular}
        }
    \end{minipage}
    \hspace{0.03\textwidth} 
    \begin{minipage}{0.39\textwidth}
        \raggedleft
        \caption{Overhead of smoothing K.}
        \label{exp:overhead_km}
        \setlength\tabcolsep{3pt}
        \scalebox{0.95}{
        \begin{tabular}{l|c|c}
        \toprule
        {\bf Model}  & {\bf Smooth K}  & {\bf TOPS $\uparrow$} \\
        \hline
        \multirow{2}{*}{\cogvideo} &  \ding{55}  & 327.57 \\
                                    & \cellcolor{gray!24}\checkmark  & \cellcolor{gray!24}\textbf{327.52} \\ \hline
        \multirow{2}{*}{\ultrapixel} & \ding{55}  & 325.18 \\
                                    & \cellcolor{gray!24}\checkmark  & \cellcolor{gray!24}\textbf{324.56} \\ 
        \bottomrule
        \end{tabular}
        }
    \end{minipage}
\end{table}

\begin{table}[h!]
\vspace{-1.5em}
    \caption{Benefit of adaptive quantization.}
    \label{exp:adaptive_quant}
    \setlength\tabcolsep{3.9pt}
    \begin{center}
    \scalebox{0.96}{
    \begin{tabular}{l||l|c|c||l|c|c}
    \toprule
    {\bf attention}  & {\bf model}  & {\bf CLIPSIM $\uparrow$} & {\bf TOPS $\uparrow$}  & {\bf Model} & {\bf MMLU $\uparrow$} & {\bf TOPS $\uparrow$}    \\ \hline
    \ourt  & \multirow{2}{*}{\cogvideo} &  0.1827  &  292.17 & \multirow{2}{*}{\llama}  & 0.46   &  208.59 \\  
    \cellcolor{gray!24}\our   &   & \cellcolor{gray!24}\textbf{0.1835}  & \cellcolor{gray!24}\textbf{327.57}  &   &  \cellcolor{gray!24}\textbf{0.46}   &  \cellcolor{gray!24}\textbf{231.74} \\ \bottomrule
    \end{tabular}
    }
    \end{center}
    \vspace{-1.2em}
\end{table}

\subsection{Ablation Study}
\vspace{-.75em}
\textbf{Overhead of smoothing K.} Table~\ref{exp:overhead_km} presents the overhead associated with smoothing K on the attention speed in real models. The results indicate a minimal reduction, less than 0.2\%.

\textbf{Benefit of adaptive quantization.} We analyzed the performance differences between using only \ourt and employing an adaptive strategy (\our). Table~\ref{exp:adaptive_quant} presents the metrics and average speed of attention on \cogvideo and \llama. The results indicate that the adaptive strategy increases the speed of attention by 11.7\% without any loss in metrics.

\vspace{-.3em}
\section{Conclusion and Future Work}
\vspace{-.6em}
We introduce \our, an efficient and precise INT8 quantization method for attention. First, we propose a method to smooth matrix K, enhancing the accuracy with under 0.2\% speed overhead. Second, we use FP16 accumulators in the Matmul of (P, V) to boost both accuracy and speed. Third, we use adaptive quantization to further improve OPS by 12\% without sacrificing accuracy. Our method surpasses FlashAttention2 and xformers by approximately \textbf{2.1x} and \textbf{2.7x}, respectively. Extensive testing confirms that our approach maintains end-to-end metrics across various models, including language, image, and video generation models. 





\section*{Acknkowlegement}
The authors would like to thank Haofeng Huang for his valuable help with the implementation. 
This work was supported by the NSFC Project (No.~62376131), Tsinghua Institute for Guo Qiang, and the High Performance Computing Center, Tsinghua University. J.Z is also supported by the XPlorer Prize.

\bibliography{main}

\begin{thebibliography}{81}
\providecommand{\natexlab}[1]{#1}
\providecommand{\url}[1]{\texttt{#1}}
\expandafter\ifx\csname urlstyle\endcsname\relax
  \providecommand{\doi}[1]{doi: #1}\else
  \providecommand{\doi}{doi: \begingroup \urlstyle{rm}\Url}\fi

\bibitem[Bao et~al.(2023)Bao, Nie, Xue, Cao, Li, Su, and Zhu]{bao2022uni}
Fan Bao, Shen Nie, Kaiwen Xue, Yue Cao, Chongxuan Li, Hang Su, and Jun Zhu.
\newblock All are worth words: A vit backbone for diffusion models.
\newblock In \emph{CVPR}, 2023.

\bibitem[Bikshandi \& Shah(2023)Bikshandi and Shah]{bikshandi2023case}
Ganesh Bikshandi and Jay Shah.
\newblock A case study in cuda kernel fusion: Implementing flashattention-2 on nvidia hopper architecture using the cutlass library.
\newblock \emph{arXiv preprint arXiv:2312.11918}, 2023.

\bibitem[Chen et~al.(2023)Chen, Qian, Tang, Lai, Liu, Han, and Jia]{chen2023longlora}
Yukang Chen, Shengju Qian, Haotian Tang, Xin Lai, Zhijian Liu, Song Han, and Jiaya Jia.
\newblock Longlora: Efficient fine-tuning of long-context large language models.
\newblock \emph{arXiv preprint arXiv:2309.12307}, 2023.

\bibitem[Choromanski et~al.(2020)Choromanski, Likhosherstov, Dohan, Song, Gane, Sarlos, Hawkins, Davis, Mohiuddin, Kaiser, et~al.]{choromanski2020rethinking}
Krzysztof Choromanski, Valerii Likhosherstov, David Dohan, Xingyou Song, Andreea Gane, Tamas Sarlos, Peter Hawkins, Jared Davis, Afroz Mohiuddin, Lukasz Kaiser, et~al.
\newblock Rethinking attention with performers.
\newblock \emph{arXiv preprint arXiv:2009.14794}, 2020.

\bibitem[Chu et~al.(2021)Chu, Tian, Wang, Zhang, Ren, Wei, Xia, and Shen]{chu2021twins}
Xiangxiang Chu, Zhi Tian, Yuqing Wang, Bo~Zhang, Haibing Ren, Xiaolin Wei, Huaxia Xia, and Chunhua Shen.
\newblock Twins: Revisiting the design of spatial attention in vision transformers.
\newblock \emph{Advances in neural information processing systems}, 34:\penalty0 9355--9366, 2021.

\bibitem[Dao(2023)]{dao2023flashattention}
Tri Dao.
\newblock Flashattention-2: Faster attention with better parallelism and work partitioning.
\newblock \emph{arXiv preprint arXiv:2307.08691}, 2023.

\bibitem[Dao et~al.(2022)Dao, Fu, Ermon, Rudra, and R{\'e}]{dao2022flashattention}
Tri Dao, Dan Fu, Stefano Ermon, Atri Rudra, and Christopher R{\'e}.
\newblock Flashattention: Fast and memory-efficient exact attention with io-awareness.
\newblock \emph{Advances in Neural Information Processing Systems}, 35:\penalty0 16344--16359, 2022.

\bibitem[Deng et~al.(2009)Deng, Dong, Socher, Li, Li, and Fei-Fei]{deng2009imagenet}
Jia Deng, Wei Dong, Richard Socher, Li-Jia Li, Kai Li, and Li~Fei-Fei.
\newblock Imagenet: A large-scale hierarchical image database.
\newblock In \emph{2009 IEEE conference on computer vision and pattern recognition}, pp.\  248--255. Ieee, 2009.

\bibitem[Dubey et~al.(2024)Dubey, Jauhri, Pandey, Kadian, Al-Dahle, Letman, Mathur, Schelten, Yang, Fan, et~al.]{llama31model}
Abhimanyu Dubey, Abhinav Jauhri, Abhinav Pandey, Abhishek Kadian, Ahmad Al-Dahle, Aiesha Letman, Akhil Mathur, Alan Schelten, Amy Yang, Angela Fan, et~al.
\newblock The llama 3 herd of models.
\newblock \emph{arXiv preprint arXiv:2407.21783}, 2024.

\bibitem[Fu et~al.(2024{\natexlab{a}})Fu, Huang, Ning, Zhang, Chen, Wu, Wang, Huang, Li, Yan, Dai, Yang, and Wang]{fu2024moa}
Tianyu Fu, Haofeng Huang, Xuefei Ning, Genghan Zhang, Boju Chen, Tianqi Wu, Hongyi Wang, Zixiao Huang, Shiyao Li, Shengen Yan, Guohao Dai, Huazhong Yang, and Yu~Wang.
\newblock Moa: Mixture of sparse attention for automatic large language model compression, 2024{\natexlab{a}}.

\bibitem[Fu et~al.(2024{\natexlab{b}})Fu, Liu, Han, Dai, Yan, Yang, Ning, and Wang]{fu2024framefusion}
Tianyu Fu, Tengxuan Liu, Qinghao Han, Guohao Dai, Shengen Yan, Huazhong Yang, Xuefei Ning, and Yu~Wang.
\newblock Framefusion: Combining similarity and importance for video token reduction on large visual language models, 2024{\natexlab{b}}.
\newblock URL \url{https://arxiv.org/abs/2501.01986}.

\bibitem[Goyal et~al.(2017)Goyal, Khot, Summers-Stay, Batra, and Parikh]{goyal2017making}
Yash Goyal, Tejas Khot, Douglas Summers-Stay, Dhruv Batra, and Devi Parikh.
\newblock Making the v in vqa matter: Elevating the role of image understanding in visual question answering.
\newblock In \emph{Proceedings of the IEEE conference on computer vision and pattern recognition}, pp.\  6904--6913, 2017.

\bibitem[Hendrycks et~al.(2020)Hendrycks, Burns, Basart, Zou, Mazeika, Song, and Steinhardt]{2020mmlu}
Dan Hendrycks, Collin Burns, Steven Basart, Andy Zou, Mantas Mazeika, Dawn Song, and Jacob Steinhardt.
\newblock Measuring massive multitask language understanding.
\newblock 2020.

\bibitem[Hendrycks et~al.(2021)Hendrycks, Basart, Mu, Kadavath, Wang, Dorundo, Desai, Zhu, Parajuli, Guo, Song, Steinhardt, and Gilmer]{hendrycks2021rendition}
Dan Hendrycks, Steven Basart, Norman Mu, Saurav Kadavath, Frank Wang, Evan Dorundo, Rahul Desai, Tyler Zhu, Samyak Parajuli, Mike Guo, Dawn Song, Jacob Steinhardt, and Justin Gilmer.
\newblock The many faces of robustness: A critical analysis of out-of-distribution generalization.
\newblock \emph{ICCV}, 2021.

\bibitem[Hessel et~al.(2021)Hessel, Holtzman, Forbes, Le~Bras, and Choi]{hessel2021clipscore}
Jack Hessel, Ari Holtzman, Maxwell Forbes, Ronan Le~Bras, and Yejin Choi.
\newblock Clipscore: A reference-free evaluation metric for image captioning.
\newblock In \emph{Proceedings of the 2021 Conference on Empirical Methods in Natural Language Processing}, pp.\  7514--7528, 2021.

\bibitem[Heusel et~al.(2017)Heusel, Ramsauer, Unterthiner, Nessler, and Hochreiter]{heusel2017gans}
Martin Heusel, Hubert Ramsauer, Thomas Unterthiner, Bernhard Nessler, and Sepp Hochreiter.
\newblock Gans trained by a two time-scale update rule converge to a local nash equilibrium.
\newblock \emph{Advances in neural information processing systems}, 30, 2017.

\bibitem[Hu et~al.(2025)Hu, Huang, Liang, Chen, Zhang, Zhu, and Chen]{hu2025quant}
Yuezhou Hu, Weiyu Huang, Zichen Liang, Chang Chen, Jintao Zhang, Jun Zhu, and Jianfei Chen.
\newblock Identifying sensitive weights via post-quantization integral.
\newblock \emph{arXiv preprint arXiv:2503.01901}, 2025.

\bibitem[Huang et~al.(2024)Huang, Hu, Jian, Zhu, and Chen]{huang2024pruninglargelanguagemodels}
Weiyu Huang, Yuezhou Hu, Guohao Jian, Jun Zhu, and Jianfei Chen.
\newblock Pruning large language models with semi-structural adaptive sparse training, 2024.
\newblock URL \url{https://arxiv.org/abs/2407.20584}.

\bibitem[Jacob et~al.(2018)Jacob, Kligys, Chen, Zhu, Tang, Howard, Adam, and Kalenichenko]{jacob2018quantization}
Benoit Jacob, Skirmantas Kligys, Bo~Chen, Menglong Zhu, Matthew Tang, Andrew Howard, Hartwig Adam, and Dmitry Kalenichenko.
\newblock Quantization and training of neural networks for efficient integer-arithmetic-only inference.
\newblock In \emph{Proceedings of the IEEE conference on computer vision and pattern recognition}, pp.\  2704--2713, 2018.

\bibitem[Jelinek et~al.(1977)Jelinek, Mercer, Bahl, and Baker]{jelinek1977perplexity}
Fred Jelinek, Robert~L Mercer, Lalit~R Bahl, and James~K Baker.
\newblock Perplexity—a measure of the difficulty of speech recognition tasks.
\newblock \emph{The Journal of the Acoustical Society of America}, 62\penalty0 (S1):\penalty0 S63--S63, 1977.

\bibitem[Jiang et~al.(2024)Jiang, Li, Zhang, Wu, Luo, Ahn, Han, Abdi, Li, Lin, et~al.]{jiang2024minference}
Huiqiang Jiang, Yucheng Li, Chengruidong Zhang, Qianhui Wu, Xufang Luo, Surin Ahn, Zhenhua Han, Amir~H Abdi, Dongsheng Li, Chin-Yew Lin, et~al.
\newblock Minference 1.0: Accelerating pre-filling for long-context llms via dynamic sparse attention.
\newblock \emph{arXiv preprint arXiv:2407.02490}, 2024.

\bibitem[Jiang et~al.()Jiang, Yan, Yao, Zhou, Chen, and Yuan]{jianghexgen}
Youhe Jiang, Ran Yan, Xiaozhe Yao, Yang Zhou, Beidi Chen, and Binhang Yuan.
\newblock Hexgen: Generative inference of large language model over heterogeneous environment.
\newblock In \emph{Forty-first International Conference on Machine Learning}.

\bibitem[Jiang et~al.(2025{\natexlab{a}})Jiang, Fu, Yao, He, Miao, Klimovic, Cui, Yuan, and Yoneki]{jiang2025demystifying}
Youhe Jiang, Fangcheng Fu, Xiaozhe Yao, Guoliang He, Xupeng Miao, Ana Klimovic, Bin Cui, Binhang Yuan, and Eiko Yoneki.
\newblock Demystifying cost-efficiency in llm serving over heterogeneous gpus.
\newblock \emph{arXiv preprint arXiv:2502.00722}, 2025{\natexlab{a}}.

\bibitem[Jiang et~al.(2025{\natexlab{b}})Jiang, Yan, and Yuan]{jianghexgen2}
Youhe Jiang, Ran Yan, and Binhang Yuan.
\newblock Hexgen-2: Disaggregated generative inference of llms in heterogeneous environment.
\newblock In \emph{International Conference on Learning Representations (ICLR)}, 2025{\natexlab{b}}.

\bibitem[Katharopoulos et~al.(2020)Katharopoulos, Vyas, Pappas, and Fleuret]{katharopoulos2020transformers}
Angelos Katharopoulos, Apoorv Vyas, Nikolaos Pappas, and Fran{\c{c}}ois Fleuret.
\newblock Transformers are rnns: Fast autoregressive transformers with linear attention.
\newblock In \emph{International conference on machine learning}, pp.\  5156--5165. PMLR, 2020.

\bibitem[Kim et~al.(2021)Kim, Gholami, Yao, Mahoney, and Keutzer]{kim2021bert}
Sehoon Kim, Amir Gholami, Zhewei Yao, Michael~W Mahoney, and Kurt Keutzer.
\newblock I-bert: Integer-only bert quantization.
\newblock In \emph{International conference on machine learning}, pp.\  5506--5518. PMLR, 2021.

\bibitem[Lefaudeux et~al.(2022)Lefaudeux, Massa, Liskovich, Xiong, Caggiano, Naren, Xu, Hu, Tintore, Zhang, Labatut, Haziza, Wehrstedt, Reizenstein, and Sizov]{xFormers2022}
Benjamin Lefaudeux, Francisco Massa, Diana Liskovich, Wenhan Xiong, Vittorio Caggiano, Sean Naren, Min Xu, Jieru Hu, Marta Tintore, Susan Zhang, Patrick Labatut, Daniel Haziza, Luca Wehrstedt, Jeremy Reizenstein, and Grigory Sizov.
\newblock xformers: A modular and hackable transformer modelling library.
\newblock \url{https://github.com/facebookresearch/xformers}, 2022.

\bibitem[Li et~al.(2023{\natexlab{a}})Li, Chen, and Zhu]{li2023memory}
Bingrui Li, Jianfei Chen, and Jun Zhu.
\newblock Memory efficient optimizers with 4-bit states.
\newblock \emph{Advances in Neural Information Processing Systems}, 36:\penalty0 15136--15171, 2023{\natexlab{a}}.

\bibitem[Li et~al.(2025)Li, Huang, Han, Zhou, Suzuki, Zhu, and Chen]{li2025optimization}
Bingrui Li, Wei Huang, Andi Han, Zhanpeng Zhou, Taiji Suzuki, Jun Zhu, and Jianfei Chen.
\newblock On the optimization and generalization of two-layer transformers with sign gradient descent.
\newblock In \emph{International Conference on Learning Representations (ICLR)}, 2025.

\bibitem[Li et~al.()Li, Wang, Peng, Song, Liu, Li, and Qiao]{liuniformer}
Kunchang Li, Yali Wang, Gao Peng, Guanglu Song, Yu~Liu, Hongsheng Li, and Yu~Qiao.
\newblock Uniformer: Unified transformer for efficient spatial-temporal representation learning.
\newblock In \emph{International Conference on Learning Representations}.

\bibitem[Li et~al.(2023{\natexlab{b}})Li, Du, Zhou, Wang, Zhao, and Wen]{li2023evaluating}
Yifan Li, Yifan Du, Kun Zhou, Jinpeng Wang, Wayne~Xin Zhao, and Ji-Rong Wen.
\newblock Evaluating object hallucination in large vision-language models.
\newblock \emph{arXiv preprint arXiv:2305.10355}, 2023{\natexlab{b}}.

\bibitem[Lin et~al.(2014)Lin, Maire, Belongie, Hays, Perona, Ramanan, Doll{\'a}r, and Zitnick]{lin2014microsoft}
Tsung-Yi Lin, Michael Maire, Serge Belongie, James Hays, Pietro Perona, Deva Ramanan, Piotr Doll{\'a}r, and C~Lawrence Zitnick.
\newblock Microsoft coco: Common objects in context.
\newblock In \emph{Computer Vision--ECCV 2014: 13th European Conference, Zurich, Switzerland, September 6-12, 2014, Proceedings, Part V 13}, pp.\  740--755. Springer, 2014.

\bibitem[Lin et~al.(2024)Lin, Tang, Yang, Zhang, Xiao, Gan, and Han]{lin2024qservew4a8kv4quantizationcodesign}
Yujun Lin, Haotian Tang, Shang Yang, Zhekai Zhang, Guangxuan Xiao, Chuang Gan, and Song Han.
\newblock Qserve: W4a8kv4 quantization and system co-design for efficient llm serving, 2024.
\newblock URL \url{https://arxiv.org/abs/2405.04532}.

\bibitem[Liu et~al.()Liu, Zaharia, and Abbeel]{liuringattention}
Hao Liu, Matei Zaharia, and Pieter Abbeel.
\newblock Ringattention with blockwise transformers for near-infinite context.
\newblock In \emph{The Twelfth International Conference on Learning Representations}.

\bibitem[Liu et~al.(2024{\natexlab{a}})Liu, Li, Li, Li, Zhang, Shen, and Lee]{liu2024llavanext}
Haotian Liu, Chunyuan Li, Yuheng Li, Bo~Li, Yuanhan Zhang, Sheng Shen, and Yong~Jae Lee.
\newblock Llava-next: Improved reasoning, ocr, and world knowledge, January 2024{\natexlab{a}}.
\newblock URL \url{https://llava-vl.github.io/blog/2024-01-30-llava-next/}.

\bibitem[Liu et~al.(2024{\natexlab{b}})Liu, Cun, Liu, Wang, Zhang, Chen, Liu, Zeng, Chan, and Shan]{liu2024evalcrafter}
Yaofang Liu, Xiaodong Cun, Xuebo Liu, Xintao Wang, Yong Zhang, Haoxin Chen, Yang Liu, Tieyong Zeng, Raymond Chan, and Ying Shan.
\newblock Evalcrafter: Benchmarking and evaluating large video generation models.
\newblock In \emph{Proceedings of the IEEE/CVF Conference on Computer Vision and Pattern Recognition}, pp.\  22139--22149, 2024{\natexlab{b}}.

\bibitem[Liu et~al.(2021)Liu, Lin, Cao, Hu, Wei, Zhang, Lin, and Guo]{liu2021swin}
Ze~Liu, Yutong Lin, Yue Cao, Han Hu, Yixuan Wei, Zheng Zhang, Stephen Lin, and Baining Guo.
\newblock Swin transformer: Hierarchical vision transformer using shifted windows.
\newblock In \emph{Proceedings of the IEEE/CVF international conference on computer vision}, pp.\  10012--10022, 2021.

\bibitem[Merity et~al.(2022)Merity, Xiong, Bradbury, and Socher]{merity2022pointer}
Stephen Merity, Caiming Xiong, James Bradbury, and Richard Socher.
\newblock Pointer sentinel mixture models.
\newblock In \emph{International Conference on Learning Representations}, 2022.

\bibitem[Milakov \& Gimelshein(2018)Milakov and Gimelshein]{milakov2018online}
Maxim Milakov and Natalia Gimelshein.
\newblock Online normalizer calculation for softmax.
\newblock \emph{arXiv preprint arXiv:1805.02867}, 2018.

\bibitem[Paperno et~al.(2016)Paperno, Kruszewski, Lazaridou, Pham, Bernardi, Pezzelle, Baroni, Boleda, and Fern{\'a}ndez]{paperno2016lambada}
Denis Paperno, Germ{\'a}n Kruszewski, Angeliki Lazaridou, Ngoc-Quan Pham, Raffaella Bernardi, Sandro Pezzelle, Marco Baroni, Gemma Boleda, and Raquel Fern{\'a}ndez.
\newblock The lambada dataset: Word prediction requiring a broad discourse context.
\newblock In \emph{Proceedings of the 54th Annual Meeting of the Association for Computational Linguistics (Volume 1: Long Papers)}, pp.\  1525--1534, 2016.

\bibitem[Peng et~al.(2023)Peng, Wu, Wei, Zhao, Yang, Liu, Xiong, Yang, Ni, Hu, et~al.]{peng2023fp8}
Houwen Peng, Kan Wu, Yixuan Wei, Guoshuai Zhao, Yuxiang Yang, Ze~Liu, Yifan Xiong, Ziyue Yang, Bolin Ni, Jingcheng Hu, et~al.
\newblock Fp8-lm: Training fp8 large language models.
\newblock \emph{arXiv preprint arXiv:2310.18313}, 2023.

\bibitem[{PyTorch Contributors}()]{pytorchBackend2024}
{PyTorch Contributors}.
\newblock Torch backend documentation.
\newblock \url{https://pytorch.org/docs/stable/backends.html#torch.backends.cuda.enable_math_sdp}.

\bibitem[Ren et~al.(2024)Ren, Li, Chen, Pei, Shao, Guo, Peng, Song, and Zhu]{ren2024ultrapixel}
Jingjing Ren, Wenbo Li, Haoyu Chen, Renjing Pei, Bin Shao, Yong Guo, Long Peng, Fenglong Song, and Lei Zhu.
\newblock Ultrapixel: Advancing ultra-high-resolution image synthesis to new peaks.
\newblock \emph{arXiv preprint arXiv:2407.02158}, 2024.

\bibitem[Salimans et~al.(2016)Salimans, Goodfellow, Zaremba, Cheung, Radford, and Chen]{salimans2016improved}
Tim Salimans, Ian Goodfellow, Wojciech Zaremba, Vicki Cheung, Alec Radford, and Xi~Chen.
\newblock Improved techniques for training gans.
\newblock \emph{Advances in neural information processing systems}, 29, 2016.

\bibitem[Shah et~al.(2024)Shah, Bikshandi, Zhang, Thakkar, Ramani, and Dao]{shah2024flashattention}
Jay Shah, Ganesh Bikshandi, Ying Zhang, Vijay Thakkar, Pradeep Ramani, and Tri Dao.
\newblock Flashattention-3: Fast and accurate attention with asynchrony and low-precision.
\newblock \emph{arXiv preprint arXiv:2407.08608}, 2024.

\bibitem[Singh et~al.(2019)Singh, Natarajan, Shah, Jiang, Chen, Batra, Parikh, and Rohrbach]{singh2019towards}
Amanpreet Singh, Vivek Natarajan, Meet Shah, Yu~Jiang, Xinlei Chen, Dhruv Batra, Devi Parikh, and Marcus Rohrbach.
\newblock Towards vqa models that can read.
\newblock In \emph{Proceedings of the IEEE/CVF conference on computer vision and pattern recognition}, pp.\  8317--8326, 2019.

\bibitem[Su et~al.(2021)Su, Lu, Pan, Murtadha, Wen, and Liu]{su2021roformer}
Jianlin Su, Yu~Lu, Shengfeng Pan, Ahmed Murtadha, Bo~Wen, and Yunfeng Liu.
\newblock Roformer: Enhanced transformer with rotary position embedding.
\newblock \emph{arXiv preprint arXiv:2104.09864}, 2021.

\bibitem[Sun et~al.(2019)Sun, Choi, Chen, Wang, Venkataramani, Srinivasan, Cui, Zhang, and Gopalakrishnan]{sun2019hybrid}
Xiao Sun, Jungwook Choi, Chia-Yu Chen, Naigang Wang, Swagath Venkataramani, Vijayalakshmi~Viji Srinivasan, Xiaodong Cui, Wei Zhang, and Kailash Gopalakrishnan.
\newblock Hybrid 8-bit floating point (hfp8) training and inference for deep neural networks.
\newblock \emph{Advances in neural information processing systems}, 32, 2019.

\bibitem[Tillet et~al.(2019)Tillet, Kung, and Cox]{openaitriton}
Philippe Tillet, H.~T. Kung, and David Cox.
\newblock Triton: an intermediate language and compiler for tiled neural network computations.
\newblock MAPL 2019, pp.\  10–19, New York, NY, USA, 2019. Association for Computing Machinery.
\newblock ISBN 9781450367196.

\bibitem[Touvron et~al.(2023)Touvron, Martin, Stone, Albert, Almahairi, Babaei, Bashlykov, Batra, Bhargava, Bhosale, et~al.]{touvron2023llama2}
Hugo Touvron, Louis Martin, Kevin Stone, Peter Albert, Amjad Almahairi, Yasmine Babaei, Nikolay Bashlykov, Soumya Batra, Prajjwal Bhargava, Shruti Bhosale, et~al.
\newblock Llama 2: Open foundation and fine-tuned chat models.
\newblock \emph{arXiv preprint arXiv:2307.09288}, 2023.

\bibitem[Vaswani(2017)]{vaswani2017attention}
A~Vaswani.
\newblock Attention is all you need.
\newblock \emph{Advances in Neural Information Processing Systems}, 2017.

\bibitem[Venkataramanan et~al.(2023)Venkataramanan, Ghodrati, Asano, Porikli, and Habibian]{venkataramanan2023skip}
Shashanka Venkataramanan, Amir Ghodrati, Yuki~M Asano, Fatih Porikli, and Amirhossein Habibian.
\newblock Skip-attention: Improving vision transformers by paying less attention.
\newblock \emph{arXiv preprint arXiv:2301.02240}, 2023.

\bibitem[Wang et~al.(2019)Wang, Ge, Lipton, and Xing]{wang2019sketch}
Haohan Wang, Songwei Ge, Zachary Lipton, and Eric~P Xing.
\newblock Learning robust global representations by penalizing local predictive power.
\newblock \emph{Advances in Neural Information Processing Systems}, 32, 2019.

\bibitem[Wang et~al.(2020)Wang, Li, Khabsa, Fang, and Ma]{wang2020linformer}
Sinong Wang, Belinda~Z Li, Madian Khabsa, Han Fang, and Hao Ma.
\newblock Linformer: Self-attention with linear complexity.
\newblock \emph{arXiv preprint arXiv:2006.04768}, 2020.

\bibitem[Wang et~al.(2025)Wang, Chen, and Zhu]{wang2024remoe}
Ziteng Wang, Jianfei Chen, and Jun Zhu.
\newblock Remoe: Fully differentiable mixture-of-experts with relu routing.
\newblock In \emph{International Conference on Learning Representations (ICLR)}, 2025.

\bibitem[Wightman(2019)]{rw2019timm}
Ross Wightman.
\newblock Pytorch image models.
\newblock \url{https://github.com/rwightman/pytorch-image-models}, 2019.

\bibitem[Wu et~al.(2023)Wu, Zhang, Liao, Chen, Hou, Wang, Sun, Yan, and Lin]{wu2023exploring}
Haoning Wu, Erli Zhang, Liang Liao, Chaofeng Chen, Jingwen Hou, Annan Wang, Wenxiu Sun, Qiong Yan, and Weisi Lin.
\newblock Exploring video quality assessment on user generated contents from aesthetic and technical perspectives.
\newblock In \emph{Proceedings of the IEEE/CVF International Conference on Computer Vision}, pp.\  20144--20154, 2023.

\bibitem[Xi et~al.(2024{\natexlab{a}})Xi, Cai, Zhu, Lu, Keutzer, Chen, and Han]{xi2024coat}
Haocheng Xi, Han Cai, Ligeng Zhu, Yao Lu, Kurt Keutzer, Jianfei Chen, and Song Han.
\newblock Coat: Compressing optimizer states and activation for memory-efficient fp8 training.
\newblock \emph{arXiv preprint arXiv:2410.19313}, 2024{\natexlab{a}}.

\bibitem[Xi et~al.(2024{\natexlab{b}})Xi, Chen, Zhao, TEH, Chen, and Zhu]{xijetfire}
Haocheng Xi, Yuxiang Chen, Kang Zhao, KAI~JUN TEH, Jianfei Chen, and Jun Zhu.
\newblock Jetfire: Efficient and accurate transformer pretraining with int8 data flow and per-block quantization.
\newblock In \emph{Forty-first International Conference on Machine Learning}, 2024{\natexlab{b}}.

\bibitem[Xi et~al.(2025)Xi, Yang, Zhao, Xu, Li, Li, Lin, Cai, Zhang, Li, et~al.]{xi2025sparse}
Haocheng Xi, Shuo Yang, Yilong Zhao, Chenfeng Xu, Muyang Li, Xiuyu Li, Yujun Lin, Han Cai, Jintao Zhang, Dacheng Li, et~al.
\newblock Sparse videogen: Accelerating video diffusion transformers with spatial-temporal sparsity.
\newblock \emph{arXiv preprint arXiv:2502.01776}, 2025.

\bibitem[Xiao et~al.(2024)Xiao, Zhang, Han, Xiao, Lin, Zhang, Liu, and Sun]{xiao2024infllm}
Chaojun Xiao, Pengle Zhang, Xu~Han, Guangxuan Xiao, Yankai Lin, Zhengyan Zhang, Zhiyuan Liu, and Maosong Sun.
\newblock Infllm: Training-free long-context extrapolation for llms with an efficient context memory.
\newblock In \emph{First Workshop on Long-Context Foundation Models@ ICML 2024}, 2024.

\bibitem[Xiao et~al.(2023{\natexlab{a}})Xiao, Lin, Seznec, Wu, Demouth, and Han]{xiao2023smoothquant}
Guangxuan Xiao, Ji~Lin, Mickael Seznec, Hao Wu, Julien Demouth, and Song Han.
\newblock Smoothquant: Accurate and efficient post-training quantization for large language models.
\newblock In \emph{International Conference on Machine Learning}, pp.\  38087--38099. PMLR, 2023{\natexlab{a}}.

\bibitem[Xiao et~al.(2023{\natexlab{b}})Xiao, Tian, Chen, Han, and Lewis]{xiao2023efficient}
Guangxuan Xiao, Yuandong Tian, Beidi Chen, Song Han, and Mike Lewis.
\newblock Efficient streaming language models with attention sinks.
\newblock \emph{arXiv preprint arXiv:2309.17453}, 2023{\natexlab{b}}.

\bibitem[Xu et~al.(2024)Xu, Liu, Wu, Tong, Li, Ding, Tang, and Dong]{xu2024imagereward}
Jiazheng Xu, Xiao Liu, Yuchen Wu, Yuxuan Tong, Qinkai Li, Ming Ding, Jie Tang, and Yuxiao Dong.
\newblock Imagereward: Learning and evaluating human preferences for text-to-image generation.
\newblock \emph{Advances in Neural Information Processing Systems}, 36, 2024.

\bibitem[Yang et~al.(2025)Yang, Xi, Zhao, Li, Zhang, Cai, Lin, Li, Xu, Peng, et~al.]{yang2025sparse}
Shuo Yang, Haocheng Xi, Yilong Zhao, Muyang Li, Jintao Zhang, Han Cai, Yujun Lin, Xiuyu Li, Chenfeng Xu, Kelly Peng, et~al.
\newblock Sparse videogen2: Accelerate video generation with sparse attention via semantic-aware permutation.
\newblock \emph{arXiv preprint arXiv:2505.18875}, 2025.

\bibitem[Yang et~al.(2024)Yang, Teng, Zheng, Ding, Huang, Xu, Yang, Hong, Zhang, Feng, et~al.]{yang2024cogvideox}
Zhuoyi Yang, Jiayan Teng, Wendi Zheng, Ming Ding, Shiyu Huang, Jiazheng Xu, Yuanming Yang, Wenyi Hong, Xiaohan Zhang, Guanyu Feng, et~al.
\newblock Cogvideox: Text-to-video diffusion models with an expert transformer.
\newblock \emph{arXiv preprint arXiv:2408.06072}, 2024.

\bibitem[Yu et~al.(2022)Yu, Luo, Zhou, Si, Zhou, Wang, Feng, and Yan]{yu2022metaformer}
Weihao Yu, Mi~Luo, Pan Zhou, Chenyang Si, Yichen Zhou, Xinchao Wang, Jiashi Feng, and Shuicheng Yan.
\newblock Metaformer is actually what you need for vision.
\newblock In \emph{Proceedings of the IEEE/CVF conference on computer vision and pattern recognition}, pp.\  10819--10829, 2022.

\bibitem[Zhang et~al.(2025{\natexlab{a}})Zhang, Huang, Zhang, Wei, Zhu, and Chen]{zhang2024sageattention2}
Jintao Zhang, Haofeng Huang, Pengle Zhang, Jia Wei, Jun Zhu, and Jianfei Chen.
\newblock Sageattention2: Efficient attention with thorough outlier smoothing and per-thread int4 quantization.
\newblock In \emph{International Conference on Machine Learning (ICML)}, 2025{\natexlab{a}}.

\bibitem[Zhang et~al.(2025{\natexlab{b}})Zhang, Huang, Zhang, Wei, Zhu, and Chen]{zhang2025sageattention2_wksp}
Jintao Zhang, Haofeng Huang, Pengle Zhang, Jia Wei, Jun Zhu, and Jianfei Chen.
\newblock Sageattention2: Efficient attention with smoothing q and per-thread quantization.
\newblock 2025{\natexlab{b}}.

\bibitem[Zhang et~al.(2025{\natexlab{c}})Zhang, Li, and Su]{sagerag}
Jintao Zhang, Guoliang Li, and Jinyang Su.
\newblock Sage: A framework of precise retrieval for rag.
\newblock In \emph{2025 IEEE 41th International Conference on Data Engineering (ICDE)}. IEEE, 2025{\natexlab{c}}.

\bibitem[Zhang et~al.(2025{\natexlab{d}})Zhang, Wei, Zhang, Xu, Huang, Wang, Jiang, Zhu, and Chen]{zhang2025sageattention3}
Jintao Zhang, Jia Wei, Pengle Zhang, Xiaoming Xu, Haofeng Huang, Haoxu Wang, Kai Jiang, Jun Zhu, and Jianfei Chen.
\newblock Sageattention3: Microscaling fp4 attention for inference and an exploration of 8-bit training.
\newblock \emph{arXiv preprint arXiv:2505.11594}, 2025{\natexlab{d}}.

\bibitem[Zhang et~al.(2025{\natexlab{e}})Zhang, Xiang, Huang, Wei, Xi, Zhu, and Chen]{zhang2025spargeattn}
Jintao Zhang, Chendong Xiang, Haofeng Huang, Jia Wei, Haocheng Xi, Jun Zhu, and Jianfei Chen.
\newblock Spargeattn: Accurate sparse attention accelerating any model inference.
\newblock \emph{arXiv preprint arXiv:2502.18137}, 2025{\natexlab{e}}.

\bibitem[Zhang et~al.(2025{\natexlab{f}})Zhang, Xiang, Huang, Wei, Xi, Zhu, and Chen]{zhang2025spargeattn_wksp}
Jintao Zhang, Chendong Xiang, Haofeng Huang, Jia Wei, Haocheng Xi, Jun Zhu, and Jianfei Chen.
\newblock Spargeattn: Training-free sparse attention accelerating any model inference.
\newblock 2025{\natexlab{f}}.

\bibitem[Zhang et~al.(2025{\natexlab{g}})Zhang, Xu, Wei, Huang, Zhang, Xiang, Zhu, and Chen]{zhang2025sageattention2++}
Jintao Zhang, Xiaoming Xu, Jia Wei, Haofeng Huang, Pengle Zhang, Chendong Xiang, Jun Zhu, and Jianfei Chen.
\newblock Sageattention2++: A more efficient implementation of sageattention2.
\newblock 2025{\natexlab{g}}.

\bibitem[Zhang et~al.(2025{\natexlab{h}})Zhang, Wei, Zhang, Zhu, and Chen]{zhang2025int8train}
Pengle Zhang, Jia Wei, Jintao Zhang, Jun Zhu, and Jianfei Chen.
\newblock Accurate int8 training through dynamic block-level fallback.
\newblock \emph{arXiv preprint arXiv:2503.08040}, 2025{\natexlab{h}}.

\bibitem[Zhao et~al.(2024{\natexlab{a}})Zhao, Fang, Liu, Wan, Soedarmadji, Li, Lin, Dai, Yan, Yang, et~al.]{zhao2024vidit}
Tianchen Zhao, Tongcheng Fang, Enshu Liu, Rui Wan, Widyadewi Soedarmadji, Shiyao Li, Zinan Lin, Guohao Dai, Shengen Yan, Huazhong Yang, et~al.
\newblock Vidit-q: Efficient and accurate quantization of diffusion transformers for image and video generation.
\newblock \emph{arXiv preprint arXiv:2406.02540}, 2024{\natexlab{a}}.

\bibitem[Zhao et~al.(2024{\natexlab{b}})Zhao, Ning, Fang, Liu, Huang, Lin, Yan, Dai, and Wang]{zhao2024mixdq}
Tianchen Zhao, Xuefei Ning, Tongcheng Fang, Enshu Liu, Guyue Huang, Zinan Lin, Shengen Yan, Guohao Dai, and Yu~Wang.
\newblock Mixdq: Memory-efficient few-step text-to-image diffusion models with metric-decoupled mixed precision quantization.
\newblock In \emph{European Conference on Computer Vision}, pp.\  285--302. Springer, 2024{\natexlab{b}}.

\bibitem[Zheng et~al.(2024{\natexlab{a}})Zheng, Chen, Mao, Liu, Zhu, and Zhang]{zheng2024masked}
Kaiwen Zheng, Yongxin Chen, Hanzi Mao, Ming-Yu Liu, Jun Zhu, and Qinsheng Zhang.
\newblock Masked diffusion models are secretly time-agnostic masked models and exploit inaccurate categorical sampling.
\newblock \emph{arXiv preprint arXiv:2409.02908}, 2024{\natexlab{a}}.

\bibitem[Zheng et~al.(2024{\natexlab{b}})Zheng, He, Chen, Bao, and Zhu]{zheng2024diffusion}
Kaiwen Zheng, Guande He, Jianfei Chen, Fan Bao, and Jun Zhu.
\newblock Diffusion bridge implicit models.
\newblock \emph{arXiv preprint arXiv:2405.15885}, 2024{\natexlab{b}}.

\bibitem[Zheng et~al.(2025)Zheng, He, Chen, Bao, and Zhu]{zheng2025elucidating}
Kaiwen Zheng, Guande He, Jianfei Chen, Fan Bao, and Jun Zhu.
\newblock Elucidating the preconditioning in consistency distillation.
\newblock \emph{arXiv preprint arXiv:2502.02922}, 2025.

\bibitem[Zheng et~al.(2024{\natexlab{c}})Zheng, Peng, Yang, Shen, Li, Liu, Zhou, Li, and You]{opensora}
Zangwei Zheng, Xiangyu Peng, Tianji Yang, Chenhui Shen, Shenggui Li, Hongxin Liu, Yukun Zhou, Tianyi Li, and Yang You.
\newblock Open-sora: Democratizing efficient video production for all, March 2024{\natexlab{c}}.
\newblock URL \url{https://github.com/hpcaitech/Open-Sora}.

\end{thebibliography}
\bibliographystyle{iclr2025_conference}

\newpage

\appendix
\section{Experimental Detail} \label{sec:exp_details}

\subsection{Environment}
We implemented our Attention kernels using OpenAI Triton~\citep{openaitriton} and conducted experiments on Ubuntu 22.04 servers. Tests on the RTX 4090 utilized a server with PCIE 5.0, a 16-core Xeon(R) 6430 CPU, and 120GB DDR4 RAM, while the RTX3090 tests employed a server with a 16-core Xeon(R) 8358P CPU and 80GB DDR4 RAM.
To reproduce our results, experiments should be conducted in the environment of torch 2.4.0+cu121, triton-nightly (version of 20240816), python 3.11, and (gcc, g++) in version 9.

\subsection{Hyper-parameters for Attention Kernels}

We use 128 for a block size of $Q$, and 64 for a block size of $K$ and $V$.
The parameters {Num\_Warps} and {Num\_Stages}, which represent the number of warp schedulers and the number of processing stages in our GPU kernels, respectively, are detailed in Table~\ref{tab:kernel_para}.

\begin{table}[!htb]
\caption{Hyper-parameters for our Attention Kernels.}
\label{tab:kernel_para}
\setlength\tabcolsep{12pt}
\centering 
    \begin{tabular}{c|c|c|c} 
    \toprule
    \textbf{HeadDim} & \textbf{Causal Mask} & \textbf{Num\_Warps} & \textbf{Num\_Stages} \\
    \hline
    \bf 64  & \bf False & 4 & 3 \\ 
    \bf 64  & \bf True  & 4 & 4 \\
    \bf 128 & \bf False & 8 & 3 \\
    \bf 128 & \bf True  & 8 & 5 \\  \bottomrule
    \end{tabular}
\end{table}

\subsection{Details of datasets and models}
We choose the first 256 annotations from the COCO 2014val dataset as the prompt set for \ultrapixel and \unidiffuser image generation.
We also used the corresponding 256 images of the 256 prompts as the ground truth images to calculate the FID and sFID.
For \cogvideo, the model is trained on long texts, so we applied an open-sora prompt set, each consisting of more than 120 words.
The specific model we used for \timm is \textit{vit\_base\_patch16\_224.augreg2\_in21k\_ft\_in1k}.

\subsection{Additional Experiments} \label{sec:append_add_exp}

\begin{table}[h!]
\caption{Comparison of SageAttention with AWQ (W4A16) on Llama2.}
\label{tab:comp_awq}
\setlength\tabcolsep{2pt}
\centering
\begin{tabular}{l|c|c|c|c}
\toprule
 & Full-Precision & SageAttention & AWQ & AWQ+SageAttention \\
\hline
\textbf{Perplexity}$\downarrow$ & 5.4721 & \textbf{5.4729} & 5.5988 & 5.5998 \\
\textbf{Speedup of Linear Computation} & 0 & 0 & 0 & 0 \\
\textbf{Speedup of Attention} & 0 & \textbf{2x} & 0 & \textbf{2x} \\
\bottomrule
\end{tabular}
\end{table}

\begin{table}[h!]
\caption{Comparison of SageAttention with Q-diffusion (W8A8) on Unidiffuser.}
\label{tab:comp_qdiffusion}
\setlength\tabcolsep{15pt}
\centering
\begin{tabular}{l|c|c|c|c}
\toprule
 & FID$\downarrow$ & sFID$\downarrow$ & CLIP$\uparrow$ & ImageReward$\uparrow$ \\
\hline
\textbf{Full Precision} & 163.33 & 145.08 & 31.52 & 0.1609 \\
\textbf{SageAttention} & \textbf{166.49} & \textbf{143.18} & \textbf{31.54} & \textbf{0.1521} \\
\textbf{Q-diffusion (W8A8)} & 395.99 & 178.56 & 18.03 & -2.273 \\
\bottomrule
\end{tabular}
\end{table}

\begin{table}[h!]
\caption{Comparison of SageAttention with VIDIT-Q on CogvideoX.}
\label{tab:comp_viditq}
\setlength\tabcolsep{1.2pt}
\centering
\begin{tabular}{l|c|c|c|c|c|c}
\toprule
 & CLIPSIM$\uparrow$ & CLIP-T$\uparrow$ & VQA-a$\uparrow$ & VQA-t$\uparrow$ & FScore$\uparrow$ & End-to-end Speedup$\uparrow$ \\
\hline
\textbf{Full Precision} & 0.1837 & 0.9976 & 68.962 & 75.925 & 3.7684 & - \\
\textbf{SageAttention} & 0.1836 & 0.9976 & 68.839 & \textbf{75.037} & 3.8339 & \textbf{34.3\%} \\
\textbf{VIDIT-Q (W8A8)} & 0.1884 & 0.9974 & 68.185 & 71.011 & 3.7342 & 22\% (theoretical maximum) \\
\bottomrule
\end{tabular}
\end{table}

\subsection{Comparison with other methods}

There are some task-specific quantization methods, such as AWQ for LLMs, Q-diffusion for text-to-image, and ViDiT-Q for text-to-video applications. 
SageAttention is orthogonal to them because those works are mainly used to quantize the linear layers. Second, AWQ is only used to compress the parameters of LLMs with no acceleration effect in computation. Q-diffusion has not reported its acceleration effect in their paper and provided codes with acceleration effect in its official repository. ViDiT-Q has not provided the codes with acceleration effect in its official repository. Nonetheless, we compare SageAttention with those works as follows.
(1) We compare the perplexity of Llama2-7B on WikiText and the speedup in the prefilling stage. The results are shown in Table~\ref{tab:comp_awq}.
We compare SageAttention with Q-diffusion (W8A8) on Unidiffuser. The results are shown in Table~\ref{tab:comp_qdiffusion}.
We compare SageAttention with VIDIT-Q on CogvideoX and the results are shown in Table~\ref{tab:comp_viditq}. Since the official repository does not provide acceleration code, we estimate a theoretical maximum: the Linear layer accounts for 24\% of Cogvideo's latency, and W8A8 offers \textbf{at most} 4x speedup for the Linear layer, resulting in a theoretical maximum end-to-end speedup of $\frac{100}{100 - 24 \times \frac{3}{4}}\% = 22\%$.

\subsection{Some Insights}
Table~\ref{tab:quant_type_analysis} shows that the metric of Llama2 remains stable with quantization. The reason is that the distribution of $Q, K$, and $V$ in the attention of Llama2-7B is relatively uniform. As a result, quantizing $Q, K$, and $V$ to INT8 or FP8 does not significantly impact the accuracy of attention. This insight inspires the idea that better control over outlier activations in models could lead to more precise quantization results.
As works like \citet{fu2024moa,zhang2024sageattention2,zhang2025sageattention2_wksp}, we believe \our can also be effectively applied to various applications related to Transformers, such as MOE systems~\citep{wang2024remoe}, linear layer quantization~\citep{hu2025quant,zhang2025int8train,zhao2024vidit}, RAG systems~\citep{sagerag}, training optimization~\citep{li2023memory,li2025optimization,huang2024pruninglargelanguagemodels,xi2024coat}, heterogeneous GPU systems~\citep{jiang2025demystifying,jianghexgen,jianghexgen2}, and diffusion models acceleration~\citep{zheng2024masked,zheng2024diffusion,zheng2025elucidating,fu2024framefusion,zhao2024mixdq,xi2025sparse,zhang2025spargeattn,zhang2025spargeattn_wksp,yang2025sparse,zhang2025sageattention3,zhang2025sageattention2++}.

\begin{table}[!htb]
\caption{SageAttention based on Torch Attention.}
\label{tab:torch+sage}
\setlength\tabcolsep{12pt}
\centering 
    \begin{tabular}{c|c|c} 
    \toprule
    \textbf{Sequence Length} & \textbf{Torch Attention} & \textbf{SageAttention based on Torch Attention} \\
    \hline
    \bf 1024  & 46 & 48 \\ 
    \bf 2048  & 42  & \textbf{55} \\
    \bf 4096 & 55 & \textbf{87} \\
    \bf 8192 & OOM  & OOM \\  \bottomrule
    \end{tabular}
\end{table}

\section{Implementation based on Torch Attention}
FlashAttention is the state-of-the-art and most commonly used standard attention; another commonly used attention is Torch attention~\citep{pytorchBackend2024}. We report the implementation speeds based on Torch in Table~\ref{tab:torch+sage}.

\begin{table}[htb!]
    \centering
    \caption{Numerical error of $Q \cdot K$ using different type of quantization.}
    \label{tab:qk_error}
    \setlength\tabcolsep{12pt}
    \begin{tabular}{c||c|c}
        \toprule
        {\bf Data Type}  & {\bf Cosine Sim}  & {\bf Relative L1}   \\ \hline
        \bf INT8   &   \textbf{99.54\%}    & \textbf{0.084}   \\ 
        E4M3   &   92.83\%    & 0.342   \\ 
        E5M2   &   77.95\%    & 0.681   \\ \bottomrule
    \end{tabular}
\end{table}

\begin{table}[htb!]
    \caption{Error of quantized attention with or without smoothed K.}
    \begin{center}
    \label{tab:append_smooth_k}
    \begin{tabular}{c|c||c|c|c}
    \toprule
    {\bf Quantization Type}  & {\bf Smoothed K}  & {\bf Cosine Sim $\uparrow$}  & {\bf Relative L1 $\downarrow$}  & {\bf RMSE $\downarrow$}
    \\ \hline
    \multirow{2}{*}{Per-token (\ourt)}   & Without   & 62.24\%  & 1.187  & 0.294 \\ 
    & \textbf{With}      & \textbf{99.47\%}  & \textbf{0.045}   & \textbf{0.031} \\ \hline
    \multirow{2}{*}{Per-block (\ourb)}   & Without   & 30.60\%  & 1.286  & 0.464 \\ 
    & \textbf{With}      & \textbf{99.31\%}  & \textbf{0.072}  & \textbf{0.035} \\ \hline
    \multirow{2}{*}{Per-tensor}  & Without   & 41.40\%  & 1.554  & 0.399 \\ 
    & \textbf{With}      & \textbf{98.06\%}  & \textbf{0.126}  & \textbf{0.059} \\ \hline
    \multicolumn{2}{c||}{FlashAttention-3 (quantized version)}  &  \textbf{26.76\%}   &  \textbf{2.5354}  &  0.5378 \\ \bottomrule
    \end{tabular}
    \end{center}
\end{table}

\begin{figure}[!htb]
    \centering
    \includegraphics[width=.96\textwidth]{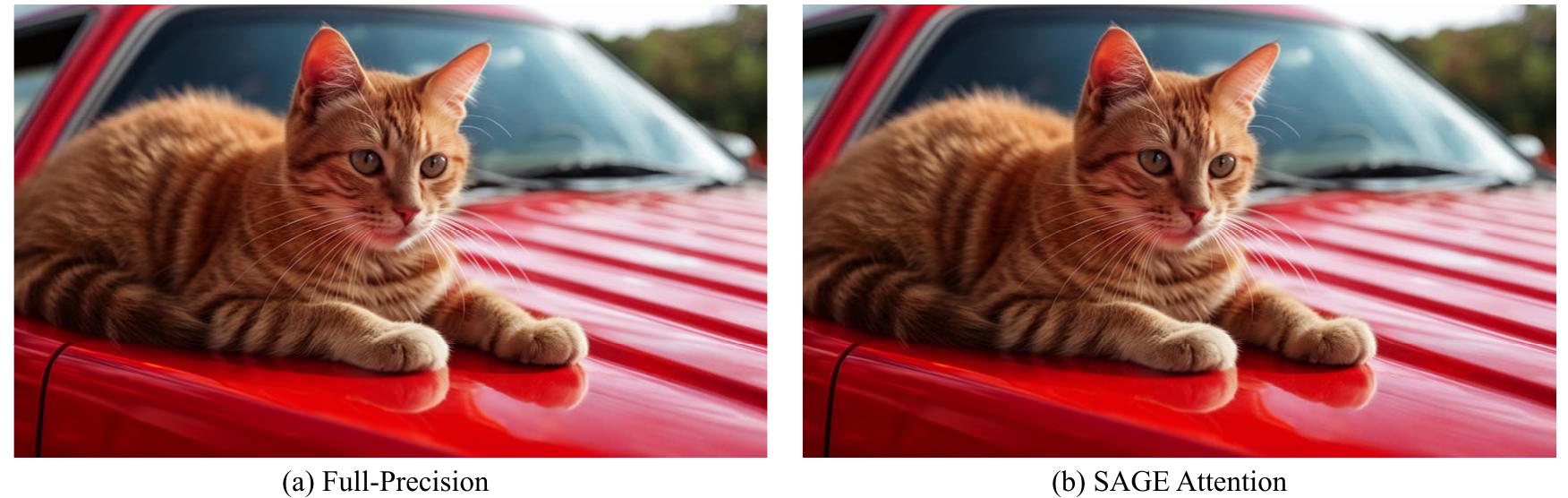}
    \caption{An image generation example of UltraPixel.}
    \label{fig:ultrapixel}
\end{figure}

\begin{figure}[!htb]
    \centering
    \includegraphics[width=.96\textwidth]{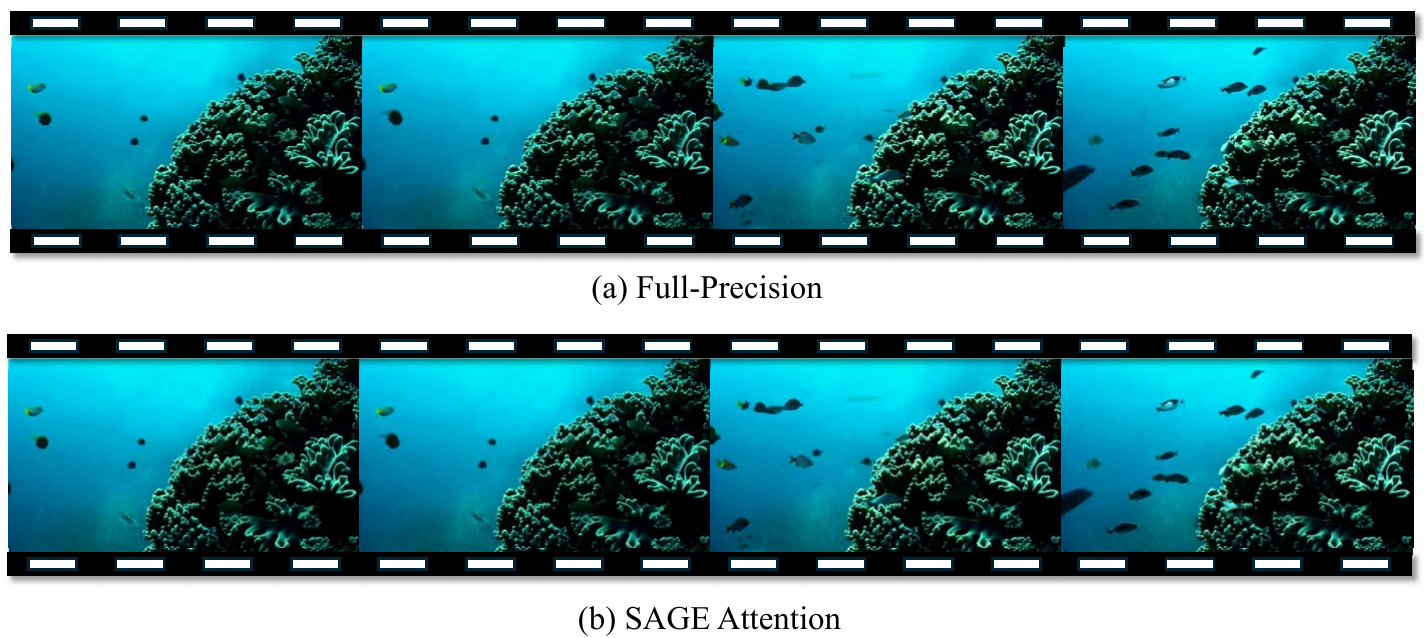}
    \caption{A video generation example of Open-Sora.}
    \label{fig:opensora}
\end{figure}

\begin{figure}[!htb]
    \centering
    \includegraphics[width=.99\textwidth]{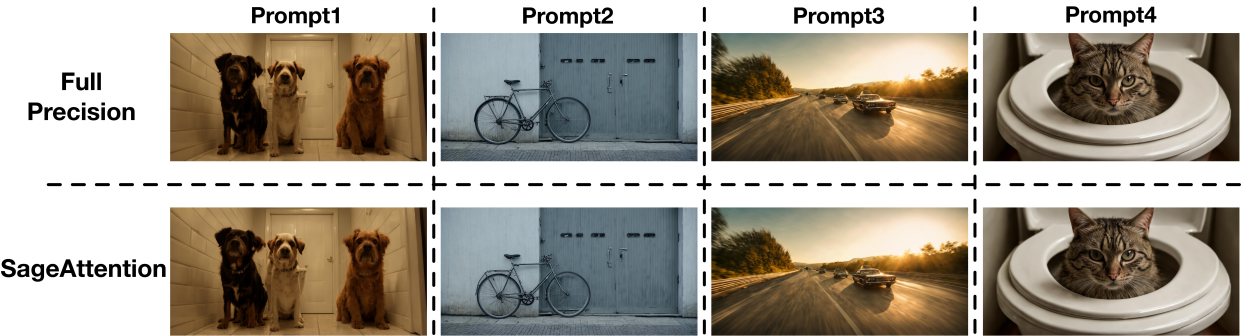}
    \caption{\purple{More image generation examples of UltraPixel, where prompt1="Two dogs are looking up while they stand near the toilet in the bathroom", prompt2="A gray bicycle is locked to some metal doors", prompt3="An image of a car driving on the highway", and prompt4="A cat on the lid of a toilet looking perturbed".}}
    \label{fig:rebuttal_image_video1}
\end{figure}

\begin{figure}[!htb]
    \centering
    \includegraphics[width=.81\textwidth]{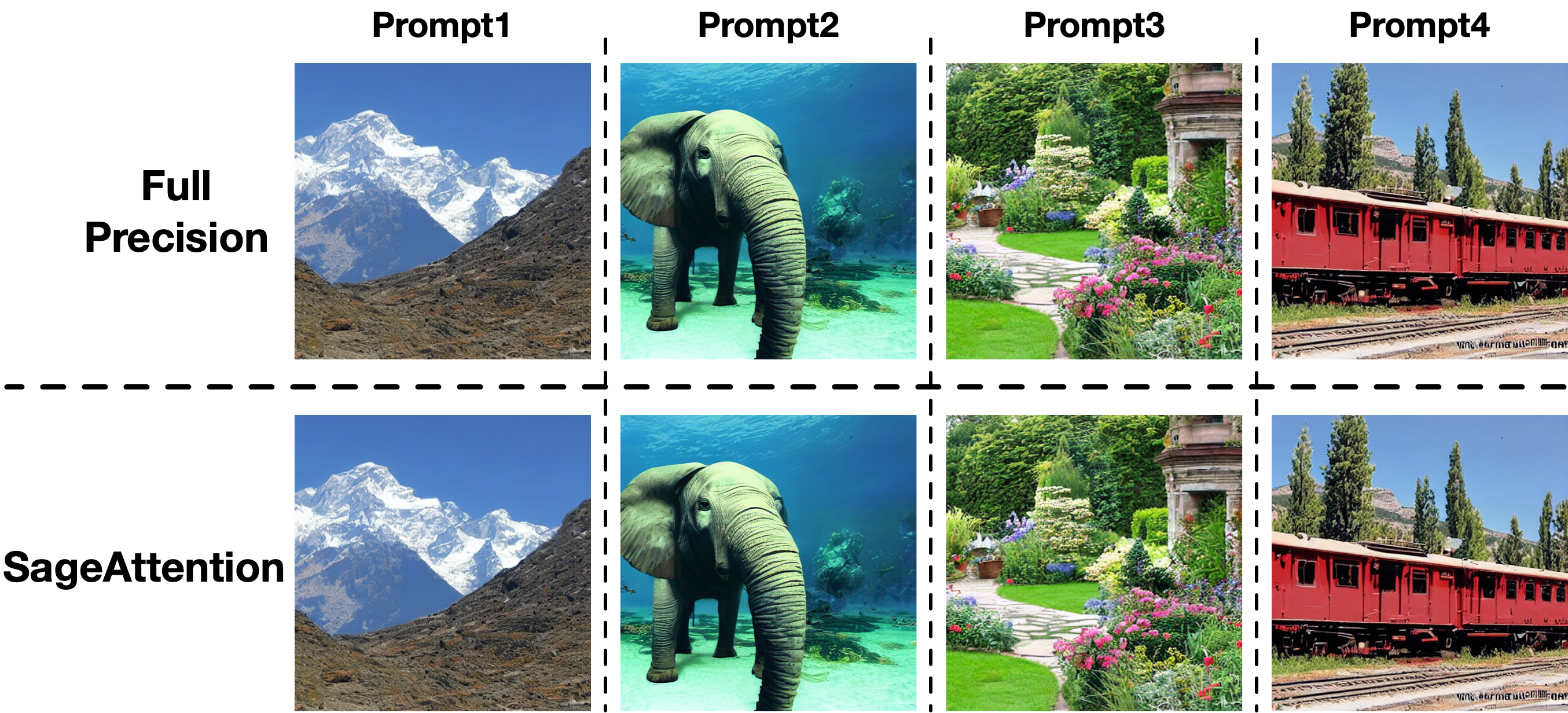}
    \caption{\purple{More image generation examples of Unidiffuser, where prompt1="Beautiful view of the Himalayas", prompt2="An elephant under the sea", prompt3="English Country Garden Design", and prompt4="An old red electric rail train in Durango, Colorado".}}
    \label{fig:rebuttal_image_video2}
\end{figure}

\begin{figure}[!htb]
    \centering
    \includegraphics[width=.999\textwidth]{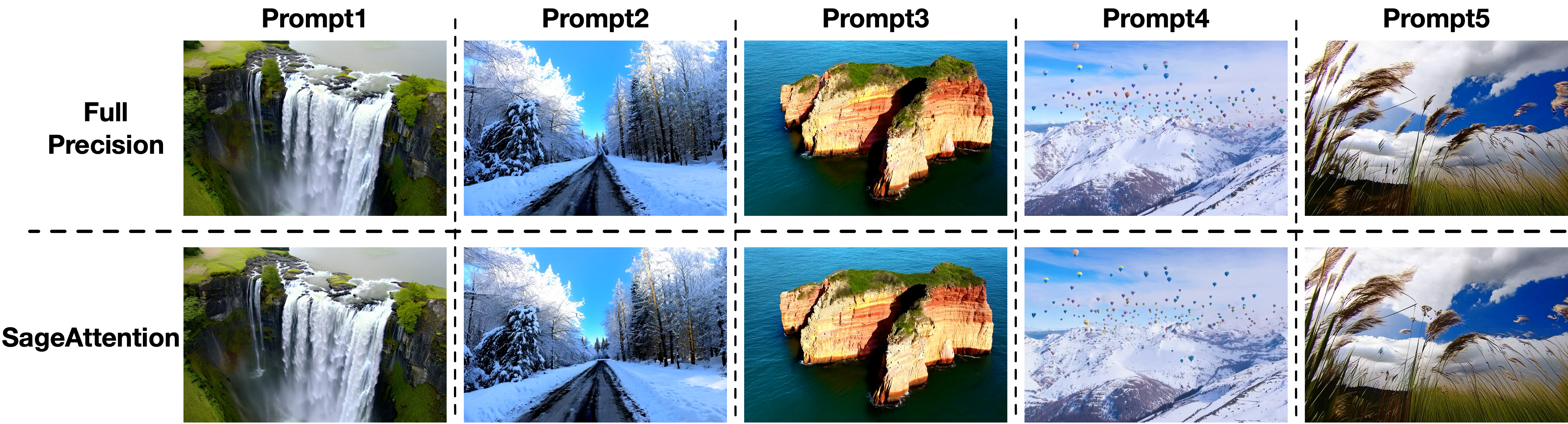}
    \caption{\purple{More image generation examples of CogvideoX. For more details about the prompts and the full videos, refer to \url{https://anonymous.4open.science/r/image_video_examples-3E44/README.md}.}
    \label{fig:rebuttal_image_video3}}
\end{figure}

\begin{table}[h!]
    \caption{Comparison of real speedup on RTX3090.}
    \label{exp:real_flops_3090}
    \begin{center}
    \begin{tabular}{l|c|l|c|c}
    \toprule
    {\bf Model}  & {\bf Shape of Q, K, V}  & {\bf Original Attention}  & {\bf SAGEAttention}  & {\bf Speedup}
    \\ \hline
    \cogvideo    & (2, 30, 17776, 64) &  71.57 (FlashAttn2)  &  \textbf{129.87}  & \cellcolor{gray!24}\textbf{1.81x}  \\
    \llama       & (4, 32, 1536, 128) &  56.54 (FlashAttn2)  &  \textbf{108.91}  & \cellcolor{gray!24}\textbf{1.93x}  \\ 
    \ultrapixel  & (2, 32, 7285, 64)  &  65.86 (FlashAttn2)  &  \textbf{131.74}  & \cellcolor{gray!24}\textbf{2.00x}  \\ 
    \unidiffuser & (4, 24, 1105, 64)  &  47.64 (xformers)    &  \textbf{108.91}  & \cellcolor{gray!24}\textbf{2.29x}  \\ 
    \timm        & (12, 64, 197, 64)  &  12.33 (Torch)        &  \textbf{66.34}  & \cellcolor{gray!24}\textbf{5.38x}  \\ \bottomrule
    \end{tabular}
    \end{center}
\end{table}

\subsection{Additional Precision Comparision}
Table~\ref{tab:qk_error} shows the precision of $Q \cdot K$ using per-token quantization in different data types compared to $Q \cdot K$ in full precision. This experiment is conducted using $Q, K$ from the 24th layer of \unidiffuser. It shows that quantizing $Q, K$ to INT8 performs higher precision than using E4M3 and E5M2.

Table~\ref{tab:append_smooth_k} shows the precision of different quantization methods with and without \textit{smoothing K} on various models. The results demonstrate that \textit{smoothing K} offers significant benefits of precision.

\subsection{Visualized Results}
Figure~\ref{fig:ultrapixel} shows the high-resolution images (2560x1536) generated by \ultrapixel using Attention of full precision and \our.
It can be seen that \our matches the full precision in the high quality and highly detailed images.
Figure~\ref{fig:opensora} shows the videos (720x1280) generated by Open-Sora~\citep{opensora} in different precisions.
\our yields identically the same video as the full precision one.

\purple{Figure~\ref{fig:rebuttal_image_video1}, Figure~\ref{fig:rebuttal_image_video2}, and Figure~\ref{fig:rebuttal_image_video3} show more visualized comparison results on \ultrapixel, \unidiffuser, and \cogvideo.}

\subsection{Real Speedup on RTX3090}

We further measure the real speed of \our and the original Attention on \unidiffuser, \ultrapixel, \cogvideo, \llama and \timm on RTX3090. 
Table~\ref{exp:real_flops} shows that \our outperforms original attention across all models. Specifically, \our yields 2.7$\times$ speedup compared to the original Attentions on average.

\end{document}